\pdfoutput=1
\documentclass[conference]{IEEEtran}
\IEEEoverridecommandlockouts
\UseRawInputEncoding
\usepackage{cite}

\usepackage{amsmath,amssymb,amsfonts}
\usepackage{graphicx}
\usepackage{textcomp}
\usepackage{xcolor}
\usepackage{color}
\usepackage{subfigure}

\usepackage{tabularx}
\usepackage{bbding}
\usepackage{booktabs}
\usepackage{multirow}
\usepackage{algorithm}
\usepackage{algpseudocode}
\usepackage{algorithmicx}
\usepackage{epstopdf}

\usepackage{soul}

\def\BibTeX{{\rm B\kern-.05em{\sc i\kern-.025em b}\kern-.08em
    T\kern-.1667em\lower.7ex\hbox{E}\kern-.125emX}}
\begin{document}

\title{Counterfactual-based minority oversampling for imbalanced classification\\
}

\author{\IEEEauthorblockN{1\textsuperscript{st} Hao Luo}
\IEEEauthorblockA{\textit{School of Big Data \& Software Engineering} \\
\textit{ChongQing University}\\
20151610@cqu.edu.cn
}
\and
\IEEEauthorblockN{2\textsuperscript{nd} Li Liu}
\IEEEauthorblockA{\textit{School of Big Data \& Software Engineering} \\
\textit{ChongQing University}\\
dcsliuli@cqu.edu.cn}
}

\maketitle

\begin{abstract}
A key challenge of oversampling in imbalanced classification is that the generation of new minority samples often neglects the usage of majority classes, resulting in most new minority sampling spreading the whole minority space.
In view of this, we present a new oversampling framework based on the counterfactual theory.
Our framework introduces a counterfactual objective by leveraging the rich inherent information of majority classes and explicitly perturbing majority samples to generate new samples in the territory of minority space.
It can be analytically shown that the new minority samples satisfy the minimum inversion, and therefore most of them locate near the decision boundary.
Empirical evaluations on benchmark datasets suggest that our approach significantly outperforms the state-of-the-art methods.
\end{abstract}


\section{Introduction}
Imbalanced classification has become an important research field, given its role in facilitating a broad range of applications using datasets with strongly imbalanced class distributions in nature, such as medical diagnosis, financial crisis prediction, network security, software defect prediction, etc.
In an imbalanced dataset, a class containing a relatively small (resp. large) number of data samples is called \emph{minority (resp. majority) class}.
Normally, the use of traditional classifiers in imbalanced domains can lead to sub-optimal classification models.
In fact, most of the existing classifiers produce inductive bias favoring the majority class in presence of an imbalanced training dataset, resulting in poor performance on other classes~\cite{Mullick2019Generative}.
For example, in cancer risk diagnosis, the cancer risk class is minority, where the number of patients diagnosed with cancer risk is much smaller than that of healthy individuals, which is majority. Unfortunately, traditional classifiers may misclassify some patients as being healthy, which is an extremely wrong release of a patient leading to severe consequences~\cite{Salina2018An}.

As the fundamental issue in data mining, many methods have been devised to tackle class imbalance over the years. We refer interested readers to a recent comprehensive review of near 70 representative imbalanced classification algorithms~\cite{Branco2016A}. Among those algorithms, the arguably most popular modeling paradigm is that of data-level methods, a.k.a.~resampling methods, which balance the data distribution over different classes by removing original data or synthesizing new data.
Those  typically fall into two main categories, i.e., undersampling which removes data from a majority class, and oversampling which generates new samples in a minority class. It is clear that undersampling discards a set of majority class samples, inevitably causing a large amount of information loss. Moreover, when the number of minority examples is small, undersampling  produces an undersized dataset, which may in turn lead to worse classification performance~\cite{Georgios2018Improving}. Hence, the undersampling approach may easily deteriorate the overall classification performance, struggling with severer data deficiency, especially for those neural network-based classifiers whose performance  increases logarithmically based on the volume of training data~\cite{Sun2017Revisiting}.

In this work, we focus on oversampling, which generates new samples of minority classes.
Intuitively, original samples can be randomly duplicate to form a new generative set. However, this set may result in overfitting in the classification model~\cite{Weiss2007Cost}.
Many oversampling approaches were proposed to avoid such an issue by generating new synthetic samples, such as SMOTE~\cite{Chawla2002SMOTE}.
Yet most of the existing approaches generate new samples primarily from the minority class data. They concentrate on the characteristics of the minority classes, which offer minor distributional information, and completely neglect the majority class.
As a result, they tend to lose a global view and generate erroneous synthetic training samples, including borderline or overlapping samples, which may have a negative impact on learnability for a classifier~\cite{Sharma2018Synthetic}.
Therefore, these approaches are rather limited in generating minority samples with meaningful information in classification problems, especially in the situation where samples from minority classes are extremely rare.

\begin{figure}[htbp]
\begin{minipage}[b]{.48\linewidth}
    \centerline{\includegraphics[width=4cm,height=3cm]{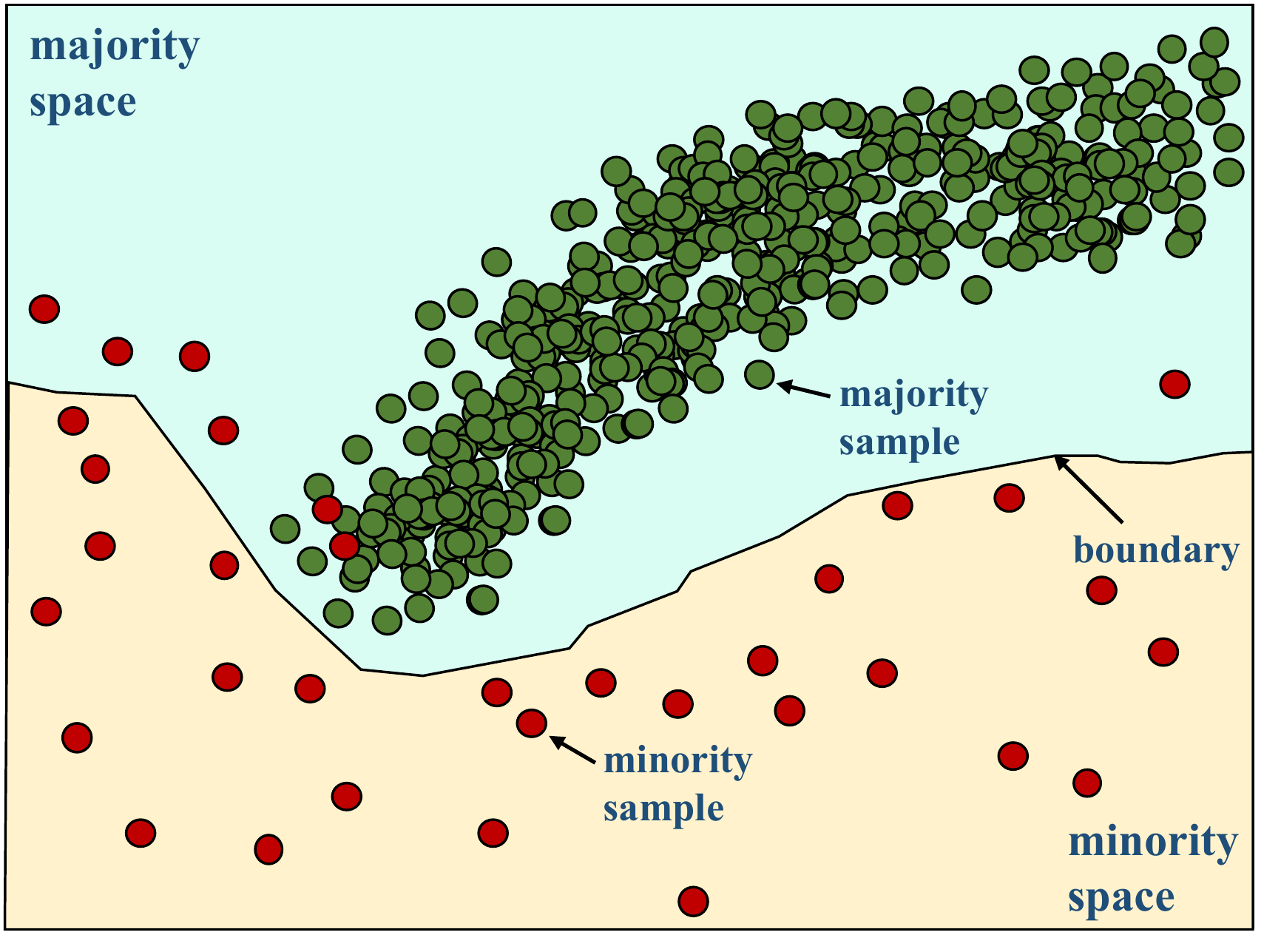}}
    \centerline{(a)}\medskip
\end{minipage}
\hfill
\begin{minipage}[b]{.48\linewidth}
    \centerline{\includegraphics[width=4cm,height=3cm]{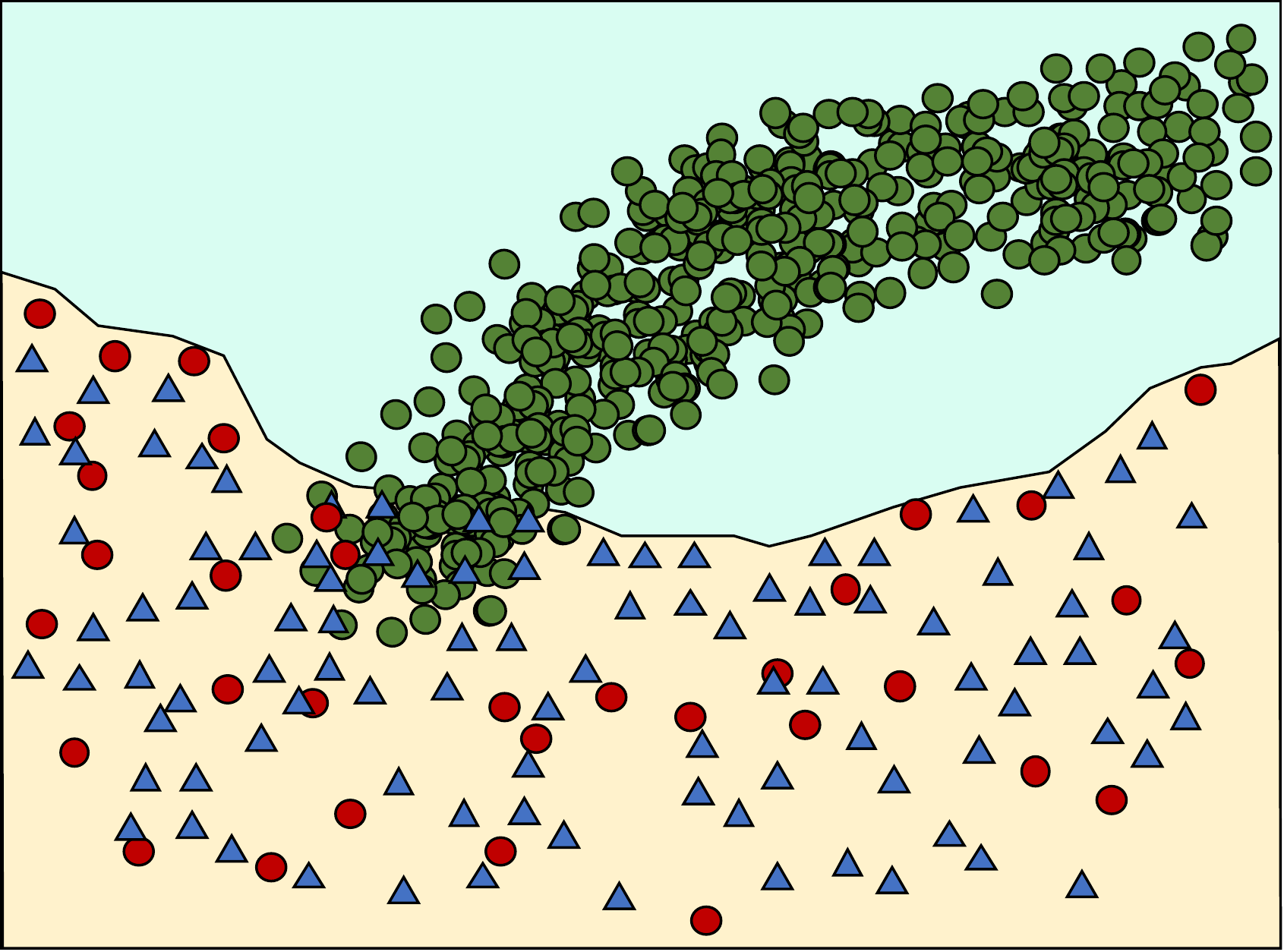}}
    \centerline{(b)}\medskip
\end{minipage}
\begin{minipage}[b]{.48\linewidth}
    \centerline{\includegraphics[width=4cm,height=3cm]{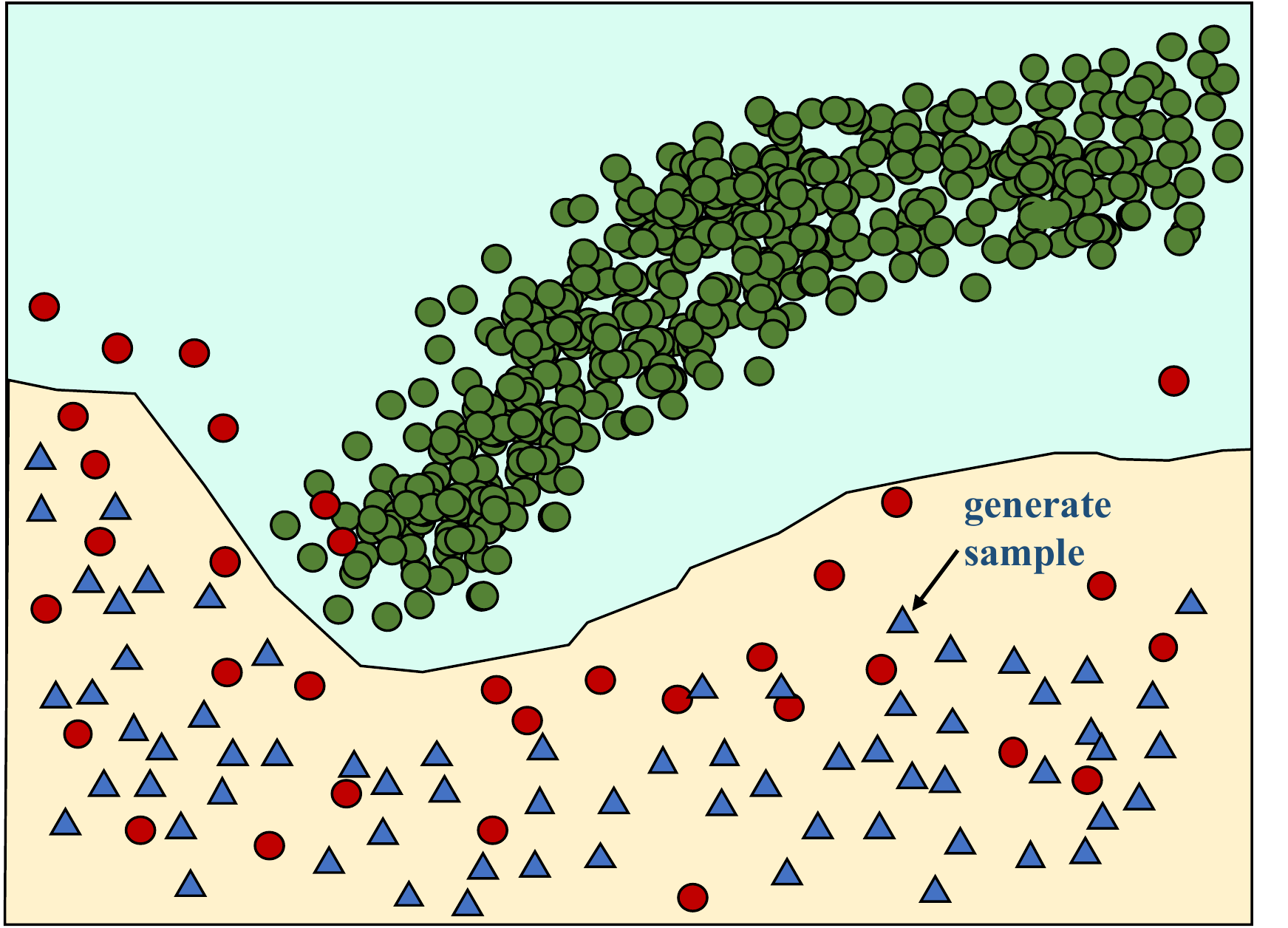}}
    \centerline{(c)}\medskip
\end{minipage}
\hfill
\begin{minipage}[b]{.48\linewidth}
    \centerline{\includegraphics[width=4cm,height=3cm]{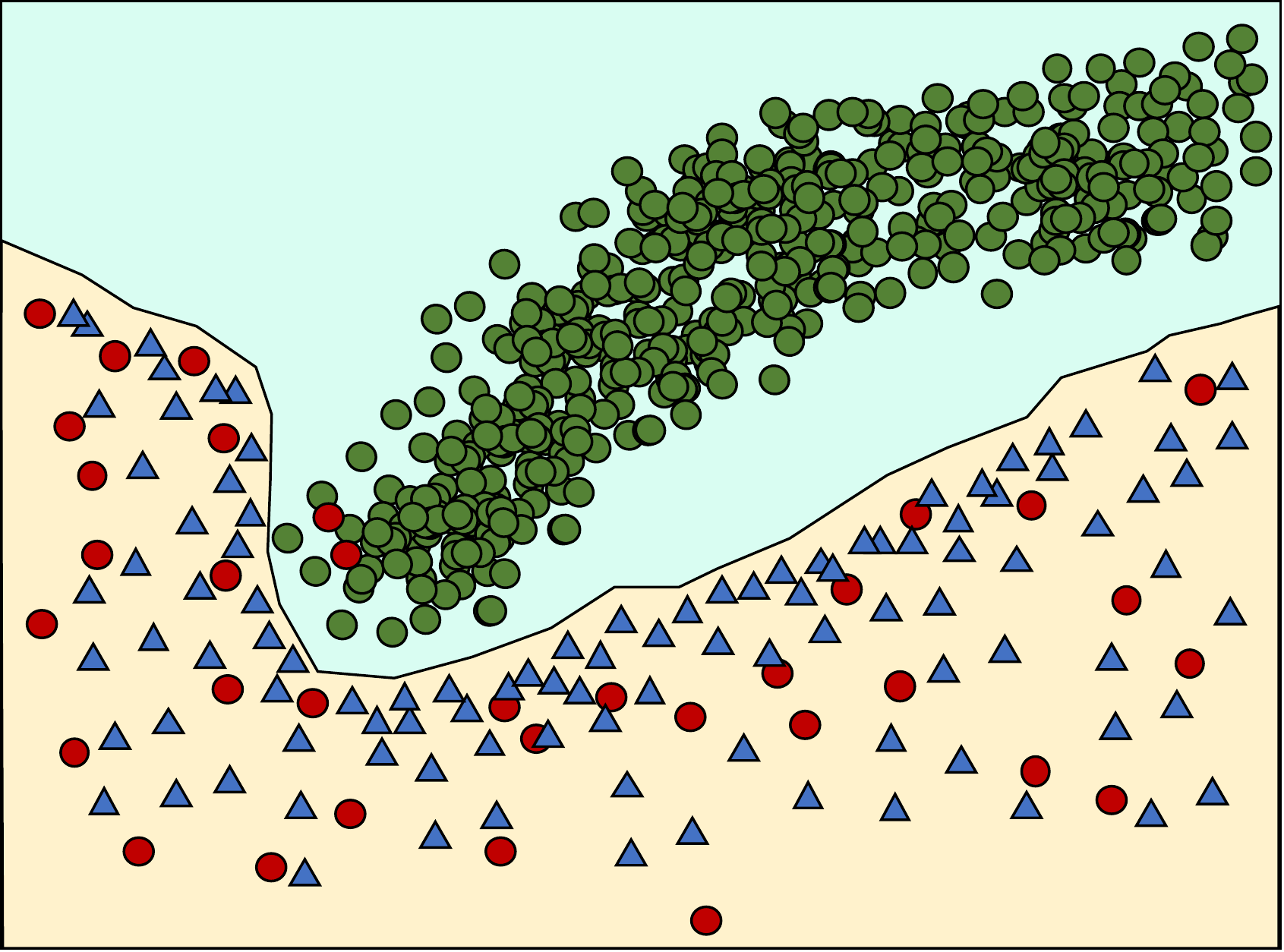}}
    \centerline{(d)}\medskip
\end{minipage}
\caption{
Illustration of decision boundary for an imbalanced binary classification. The green dots represent majority samples, while the red dots represent minority samples. An oversampling approach generates synthetic minority samples denoted by blue triangles. The decision boundary (black line) is learnt by a classifier to divide the feature space into two parts: majority space (green area) and minority space (yellow area).
(a) The classifier is trained without any generative samples but merely with original samples. It can be seen that the decision boundary is biased toward minority class, and the majority space is larger than minority space. A considerable set of minority samples are contained in the majority space, which would be detected as noise points by this classifier. As a result, it may cause a misclassification of minority samples.
(b) Synthetic minority samples are generated equally spreading out of the whole minority space, including both boundary samples and the samples far from the decision boundary. The minority space is enlarged, and thus it helps the classifier to improve the minority detection. However, the minority samples bleed into the majority space, which will negatively affect the performance of the majority class.
(c) New minority samples are generated by an approach such as SMOTE which oversamples most of them within the minority territory but far from the boundary between minority and majority classes. These generative samples would have limited contribution to the learning of decision boundary.
(d) This is an ideal performance on both classes that an oversampling approach generates most of the samples that are near the boundary. In this way, it can obtain a decision boundary that optimally divides the two classes and limits bleeding one class into the other space. To achieve this, we introduce a counterfactual-based oversampling approach in this work that generates abundant boundary samples of minority class by leveraging the rich information inherent in the majority class.
}
\label{fig:boundray}
\end{figure}

On the other hand, most classifiers learn to decide the boundary of each class as exactly as possible in order to achieve outstanding classification performance.
In fact, a better \emph{decision boundary} is as important as learning better features for a classifier.
Unfortunately, it is found that the decision boundary is biased toward minority class~\cite{Byungju2020Adjusting}. It implies that a smaller volume of the feature space is allocated to the minority class compared with the majority class.
As a result, a trained classifier is more generalized to a majority class, whereas it is often overfitted to a minority class.
The reduction of model capacity is the most representative method for resolving the overfitting problem.
However, it will deteriorate the performance of the majority class, and thus is not practical for imbalanced classification.
To this end, a large margin around the decision boundary is required for minority class to provide favorable generalization capability for a classifier.
Besides, the contribution of samples near the decision boundary (called \emph{boundary samples}) to classification model is much greater than that of other samples.
This is because these boundary samples are more likely to be misclassified than samples that are far from it. Correctly classifying boundary samples is critical for classification models.
In addition, the minority samples that are closer to the majority in the feature space are more likely to be treated as noise points. To illustrate this clearly, Fig.~\ref{fig:boundray} shows an example of an imbalanced binary classification scenario with various decision boundaries, which are trained by the same classification model but over different sets of generative minority samples.

To address these aforementioned issues in minority sample generation, we present a counterfactual-based oversampling framework for imbalanced classification.
In particular, our approach considers a principled way of generating minority samples inspired by \emph{counterfactual thinking} that explicitly leverages the rich information inherent in a majority class.
Briefly speaking, to synthesize new samples of minority class, we propose to introduce a counterfactual objective by perturbing majority samples with \emph{minimum inversion}.
The newly generated samples, called \emph{counterfactual samples}, locate in the territory of minority space and mostly near the decision boundary.
In addition, a \emph{truncated normal distribution} is introduced to ensure a high probability of generating the boundary sample close to its corresponding majority sample, which satisfies the requirement of minimum inversion.
Comparing to a variety of state-of-the-art methods, our oversampling approach is more capable of generating new boundary minority samples from existing majority samples. In our empirical evaluations (to be detailed in Section V), we demonstrate the advantage of our approach in four benchmark datasets and for some most common classifiers.

\section{Related work}

There has been a fair amount of work on imbalanced classification, which can be roughly divided into four categories: data-level methods, algorithm-level methods, cost-sensitive methods and classifier ensembles.
The algorithm-level approaches aim to modify existing models or propose specific frameworks to apply to an imbalanced domain. However, such modification requires deep knowledge about the model itself, leading to poor generaliability. Cost-sensitive method and classifier ensembles deal with imbalance by calculating the cost matrix and using multiple classifiers with a bagging combination approach, respectively. They may also bring overfitting problems to minority classes as a trained classifier is more generalized to a majority class. We refer the interested readers to the excellent reviews (Branco et al.~\cite{Branco2016A}) for these methods.

Our focus in this paper is at the data level of balancing the number of samples among classes by resampling, which further have two main paradigms: oversampling and undersampling.
The undersampling method~\cite{Lin2017Clustering,
Jianjun2019Undersampling} performs sample selection and deletion on majority class to construct a balanced dataset. After a large number of samples are deleted, it will not only cause information loss, but may also challenge machine learning algorithms due to insufficient data information.

In contrast to undersampling, oversampling generates minority samples in the imbalanced dataset.
The most famous oversampling methods are SMOTE~\cite{Chawla2002SMOTE} and its variants~\cite{
Hu2013A,lee2017gaussian-based,Ma2017CURE, Prusty2017Weighted} that generate new samples by interpolation between the nearest neighbors of minority samples.
However, most of these methods primarily rely on the minority class data during the generation, leading to the overfitting of a classifier to a minority class.
Sharma et al.~\cite{Sharma2018Synthetic} (SWIM) used the majority samples to synthesize minority class data.
However, they ignored the significance of boundary samples.
Recently, the maturity and prevalence of deep learning offer advanced capabilities to incorporate deep oversampling framework (DOS)~\cite{Ando2017Deep}.
Generative adversarial networks were often utilized as the framework to supplement minority samples in imbalanced datasets, such as text-GAN~\cite{Ankesh2018Phishing} and GAMO~\cite{Mullick2019Generative}.
However, compared with the conventional methods like SMOTE, the deep oversampling methods are computationally exhaustive. The generated samples cannot be reused if a new class arrives. Moreover, it is hard to interpret its generation process, which limits the capability of yielding boundary samples in a manageable and stable way.
We summarize the commonly used oversampling methods in Table~\ref{tab-relatework}.
This motivates us to present a new counterfactual-based oversampling method to explicitly leverage the rich information in majority classes, which is achieved by perturbing majority samples to generate minority samples near the decision boundary.
\begin{table}
\centering
\caption{A summary comparison of oversampling methods for imbalanced classification}\label{tab-relatework}
\scalebox{0.83}{
\begin{tabular}{m{0.02\textwidth}<{\centering}m{0.16\textwidth}<{\centering}m{0.07\textwidth}<{\centering}m{0.1\textwidth}<{\centering}m{0.12\textwidth}<{\centering}}
\hline
\toprule
Ref. & Methods & Synthesize with & Boundary enhancement & Classification evaluated in ref.  \\
\midrule
~\cite{Sharma2018Synthetic} & SWIM & Majority& \XSolid& Binary\\
~\cite{Rong2014Stochastic} & SSO & Minority & \Checkmark &Binary  \\
~\cite{He2008ADASYN}  & ADASYN & Minority & \Checkmark &Binary \\
~\cite{article2}  & ANS & Minority & \XSolid &Binary \\
~\cite{Georgios2018Improving}  & K-means & Minority & \XSolid &Binary \\
~\cite{koziarski2017ccr}  & CCR & Minority & \Checkmark &Binary \\
~\cite{lee2017gaussian-based} & Gassian\_SMOTE & Minority & \XSolid &Binary \\
~\cite{abdi2015combat} & MDO & Minority & \XSolid &Multiple \\
~\cite{rivera2017noise} & NRAS &Minority & \XSolid &Binary \\
~\cite{torres2016smote} & SMOTE\_D &Minority & \XSolid &Binary \\
~\cite{Ma2017CURE}  & CURE-SMOTE & Minority & \XSolid &Multiple  \\
~\cite{rivera2016priori} & OUPS &Minority & \XSolid &Binary\\
~\cite{Hu2013A} & NRSBoundary-SMOTE & Minority & \Checkmark & Binary\\
-& Ours & Majority & \Checkmark & Multiple\\
\bottomrule
\end{tabular}
}
\end{table}

\section{Background on Counterfactual and Problem Formulation}

Given a dataset $\mathbf{X}=\{(\mathbf{x},y)\}$ of $N$ samples from a set of $C$ classes, with each sample $\mathbf{x}$ ($\mathbf{x}\in \mathbb{R}^M$) consisting of $M$ observed features and being associated with a class label $y$ ($y\in\{1,\ldots,C\}$).
For any class $c$ ($1\leq c\leq C$), denote $\mathbf{X}_c\subseteq \mathbf{X}$ the corresponding subset of $N_c$ samples. Here each sample $\mathbf{x}\in \mathbf{X}_c$ is an instance of the $c$-th class.
Without loss of generality, we assume the classes to be ordered such that $N_1\leq N_2\leq \ldots \leq N_C$. Furthermore, we have $P_1\leq P_2\leq \ldots \leq P_C$, where $P_c$ is the prior probability of the $c$-th class.
For any pair of classes $i,j$ ($1\leq i<j\leq C$) such that $N_i<N_j$, we say $\mathbf{X}_i$ is the minority class, while $\mathbf{X}_j$ is the majority class.
In this work we aim to generate new samples of the minority class $i$ by leveraging the rich information inherent in the majority class $j$ inspired by the counterfactual thinking.

The term \emph{counterfactual} is defined by the Merriam-Webster dictionary as contrary to the facts.
In psychology, the concept of counterfactual involves the human tendency to create possible alternatives to events that have already occurred or something that is contrary to what actually existed.
A counterfactual account of causation also exists in philosophy, which began with the seminal 1973 article of David Lewis, titled ``Causation''~\cite{lewis1974causation}.
The account employs counterfactual conditionals that are interpreted as statements of consequences of \emph{hypothetical interventions}~\cite{Zhang2013A}.
For example, given that an individual was classified into the group of no risk of lung cancer (i.e. majority class), and given that he has no history of smoking (i.e. one of the observed features), and given everything else we know about the circumstances of the person (i.e. other observed features), what can we say about the probability of the individual having the risk of lung cancer, had he had a long history of smoking?
The implication of this example is that if we know all the observed features about an event and the result of that event, and now we intervene on some of them. Can we quantify the impact of the intervention on the result, or can we obtain a contrary result (e.g. high risk of lung cancer, the minority class)?

Judea Pearl~\cite{Pearl2009Causality} provided an elegant formal semantics of such intervention based on functional causal models for the counterfactuals, defined as \emph{do-calculus} $do(\mathbf{x})$.
In our case, $\mathbf{x}=X$ is a sample in the dataset $\mathbf{X}$, called \emph{factual sample}, where $X$ is its original (factual) value. Its corresponding class label $y$ is called \emph{factual label}.
The do-calculus simulates physical intervention by changing the value of the factual sample $\mathbf{x}$ to $X'$, which can also be viewed as the generation of a new sample $\mathbf{x}' = X'$, called \emph{counterfactual sample}, while keeping the rest of the samples in the factual dataset $\mathbf{X}$ unchanged.
Here the superscript prime $'$ means the counterfactual after the intervention.
The postintervention distribution resulting from the operation $do(\mathbf{x} = X')$ is given by
\begin{equation}
\begin{aligned}
p(&y'|do(\mathbf{x}\!=\!X'))\!=p(y'|\mathbf{x'}\!=\!X')\!\\
&=\int_{\mathbf{X}}p\left(y'|\mathbf{x}'\!=\!X',\mathbf{X}=\dot{X}\right)p(\mathbf{X})d\mathbf{X}\\
&=\mathbb{E}_{p_{\mathbf{X}}} p\left(y' | \mathbf{x}'\!=\!X',\mathbf{X}=\dot{X}\right),
\end{aligned}
\label{eq:do}
\end{equation}
where $y'$ is the class label associated with the new counterfactual sample $\mathbf{x}'$ after the \emph{do} operation, called \emph{counterfactual label}. $\dot{X}$ denotes the factual values of all the samples in the dataset $\mathbf{X}$.

The probability $p\left(y'|\mathbf{x}'\!=\!X'\right)$ in Eq.(\ref{eq:do}) is the joint distribution of the data observed after the sample $\mathbf{x}$ is forced to intervene. Actually, such conditional probability of intervention is the expected values of counterfactuals of the observable factual samples.
Unlike the joint distribution of overall data observed in intervention, counterfactual method focuses on individuals. Because the specificity of each sample, we are not interested in the expected value of the observed distribution, but only in calculating the probability of each individual.
Looking specifically at each sample, counterfactual method constructs its most similar parallel twin points for each sample (i.e. factual and counterfactual) through intervention. For example, considering a dataset $\mathbf{X}$ of three samples, $\mathbf{X}=\{\mathbf{x}_1,\mathbf{x}_2,\mathbf{x}_3\}$, if the factual sample $\mathbf{x}_1$ is intervened by a new counterfactual sample $\mathbf{x}_1'$, i.e. $\{\mathbf{x}_1',\mathbf{x}_2,\mathbf{x}_3\}$, the counterfactual in our work aims to estimate the following conditional probability:
\begin{equation}
\begin{aligned}
&p\left(y' | \mathbf{x}'\!=\!X',\mathbf{X}=\dot{X}\right)\\
&=p\left(y' | \mathbf{x}_1'\!=\!X_1',\mathbf{x}_1\!=\!X_1,\mathbf{x}_2\!=\!X_2, \mathbf{x}_3\!=\!X_3\right).
\end{aligned}
\label{eq:counterfatual}
\end{equation}

\begin{figure}[htbp]
\centerline{\includegraphics[width=0.6\columnwidth]{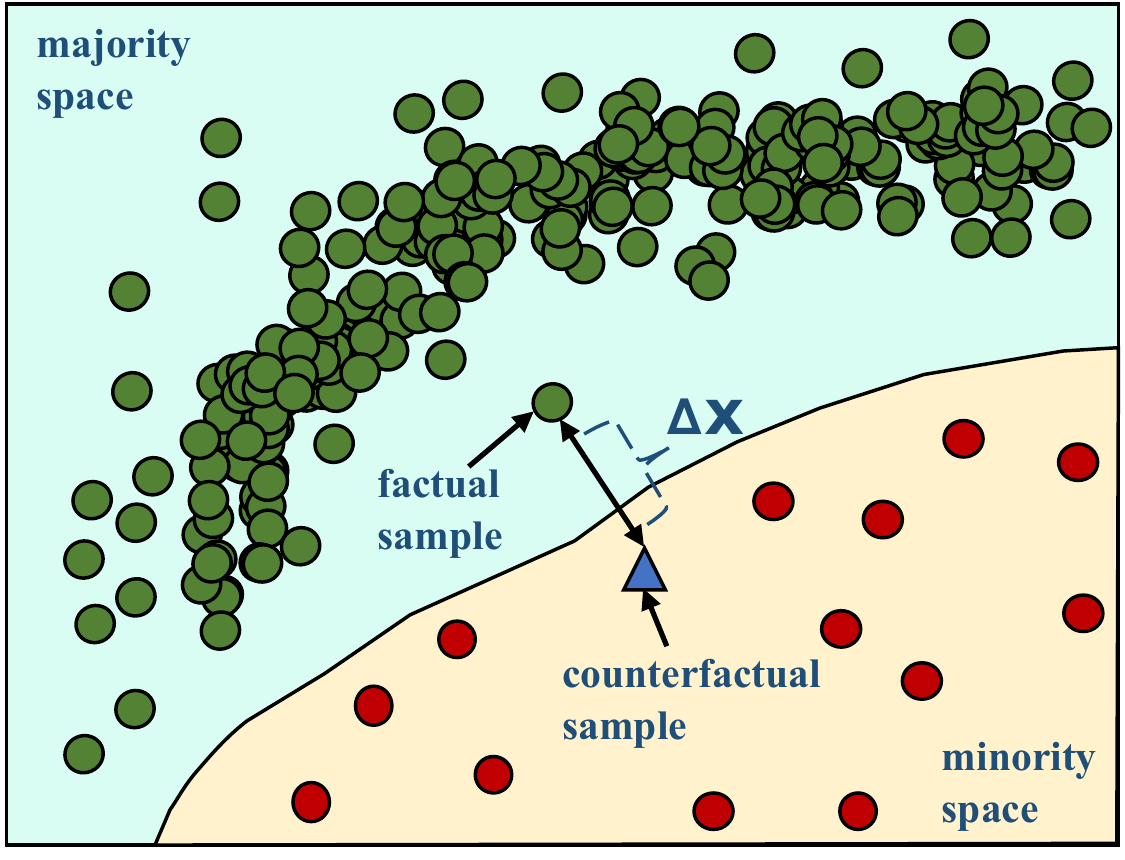}}
\caption{An illustration of a counterfactual sample in imbalanced classification.}
\label{fig:couterfactual_sample}
\end{figure}

Practically in the imbalanced classification problem, a counterfactual sample can be generated by intervening on one of its features and keep the remaining features unchanged. The core of counterfactual method in our work is to perform minimal intervention on the sample feature to change the class label of the sample (i.e., ``minimum inversion'').
In other words, a new counterfactual sample is required to be as similar as possible to its corresponding factual sample.
Formally, given a classification model $y=f_\mathbf{w}(\mathbf{x})$, the counterfactual of a factual sample $\mathbf{x}$ with its factual label $y$ (i.e. $y=f_\mathbf{w}(\mathbf{x})$) can be generated by a small perturbation $\Delta x$ such that $y'=f_\mathbf{w}(\mathbf{x}+\Delta \mathbf{x})$, where $y'\neq y$.
For example, in an imbalanced binary classification, given a factual sample $\mathbf{x}\in \mathbf{X}_2$ in majority class (i.e. $y=2$), a counterfactual sample of $\mathbf{x}$ is $\mathbf{x}+\Delta \mathbf{x}$ such that $y'=1$. That is to say, the new generative counterfactual sample belongs to minority class, as illustrated in Fig.~\ref{fig:couterfactual_sample}.
It is worth to mention that the classification model $f_\mathbf{w}$ is trained based on the factual dataset $\mathbf{X}$ and it will remain unchanged during the generation of counterfactual samples of minority classes.

In this way, it is possible to generate new samples of the minority class by perturbing the samples of the majority class.
While intervening, one of the rules required by counterfactual theory has to be followed: The counterfactual sample should be as close as possible to the factual sample. That is to say, a minimum perturbation $\Delta x$ on a factual sample $\mathbf{x}$ of the majority class is required to ensure the inversion of the classification label to the minority class. Generally, a distance metric is often recommended to characterize this requirement.
As a result, most of the new counterfactual samples will be concentrated near the decision boundary, due to the closeness to their corresponding factual samples.
This inspires us to present in what follows a counterfactual-based oversampling method where these minority samples can be systematically generated from the majority class by implementing interventions with minimum perturbation.

\section{Our approach}
\begin{figure*}[!htbp]
\centerline{\includegraphics[width=18cm]{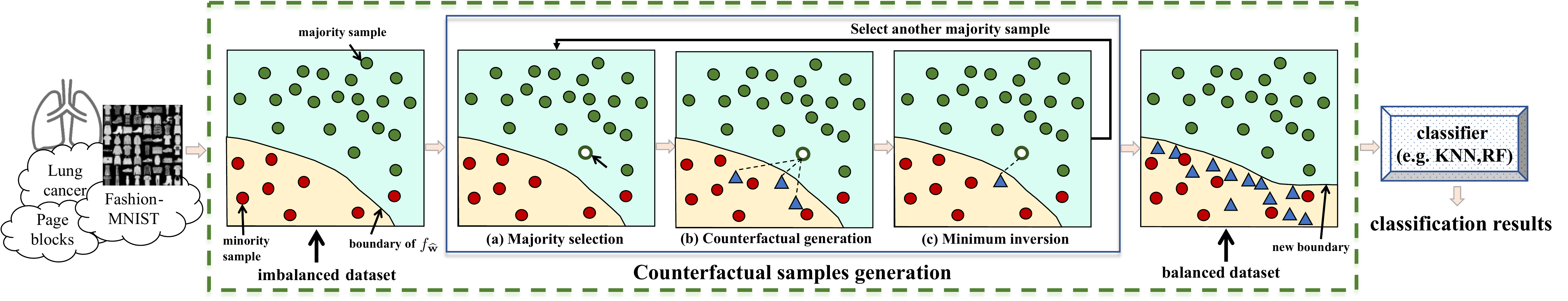}}
\caption{\small{The framework of our approach.}}
\label{fig:framework}
\end{figure*}

Let us consider a dataset $\mathbf{X}$ of $N$ samples $\{(\mathbf{x}_n,y_n)\}$ over $C$ classes, $n=1,\ldots,N$, where $y_n$ is the class label of the $n$-th sample $\mathbf{x}_n$.
Here, each sample $\mathbf{x}_n$ contains $M$ features, denoted by $\mathbf{x}_n=[X_{n1},\ldots,X_{nM}]$, where $X_{nm}$, $m=1,\ldots, M$, is an instance of the $m$-th feature variable of the $n$-th sample.
For any pair of classes $i,j$ where $N_i<N_j$, our objective is to generate a set of new counterfactual samples of the minority class $i$ by perturbing the factual samples in the majority class $j$ in an efficient way. The main procedure of our approach is illustrated in Fig.~\ref{fig:framework}.

\subsection{Counterfactual goal}
To generate counterfactual samples, we first need to determine a classifier $f_\mathbf{w}$ trained by the factual samples in the dataset $\mathbf{X}$.
To maximize the number of correctly classified samples, the optimal parameter vector $\widehat{\mathbf{w}}$ of the classifier $f_\mathbf{w}$ over dataset $\mathbf{X}$ can be calculated as
\begin{equation}
\begin{aligned}
\widehat{\mathbf{w}}
& = \arg \min _{\mathbf{w}}\mathcal{L}_f(\mathbf{w},\mathbf{X})\\
& =\arg \min _{\mathbf{w}} \frac{1}{N}\sum_{n=1}^{N} \ell\left(f_\mathbf{w}\left(\mathbf{x}_{n}\right), y_{n}\right)+\eta(\mathbf{w}),
\end{aligned}
\end{equation}
where $\mathcal{L}_f$ is the loss function with a regularization penalty $\eta(\mathbf{w})$.
In this work we use the common squared error loss, i.e. $\ell(f_\mathbf{w}(\mathbf{x}_{n}), y_n)=(f_\mathbf{w}(\mathbf{x}_n)-y_n)^2$, and the L2-norm penalty $\eta(\mathbf{w})=\rho\sum_{w\in\mathbf{w}}w^2$, where $\rho$ is a term-regularization parameter to avoid overfitting.

Given a factual sample $\mathbf{x}_n$ of the majority class $j$ (i.e. $y_n=j$), a counterfactual sample $\mathbf{x}'_n$ is required to be as close to it as possible, while it belongs to the minority class $i$ ($N_i<N_j$), i.e. $y'_n=i$.
Thus, for any pair of classes $i,j$ ($N_i<N_j$), each factual sample $\mathbf{x}_n\in \mathbf{X}_j$ is used to generate a corresponding counterfactual sample $\mathbf{x}'_n$ by holding $f_{\widehat{\mathbf{w}}}$ fixed and minimizing the objective as follows
\begin{equation}
\mathcal{L}\left(\mathbf{x}'_n, y'_{n}, \mathbf{x}_{n}, \lambda\right)=\lambda \cdot\left(f_{\widehat{\mathbf{w}}}\left(\mathbf{x}'_{n}\right)-y'_{n}\right)^{2}+d\left(\mathbf{x}_{n}, \mathbf{x}'_{n}\right),
\end{equation}
where $d\left(\mathbf{x}_{n}, \mathbf{x}'_{n}\right)$ is the distance metric to measure the closeness between $\mathbf{x}_n$ and $\mathbf{x}'_n$, and $\lambda$ is a tuning parameter, which controls the generation of the counterfactual samples.
A large $\lambda$ can guarantee to generate counterfactual samples relatively more flexible than that of a small $\lambda$, while a small $d\left(\mathbf{x}_{n}, \mathbf{x}'_{n}\right)$ ensures that the new counterfactual sample locates near its corresponding factual sample as close as possible.

Our goal is to find an optimal counterfactual sample $\mathbf{x}'_n$ for a given $\mathbf{x}_n$, which is expected to fall into the class $i$ (i.e., $y'=i$). This is formulated by the following objective
\begin{equation}
\begin{aligned}
&\widehat{\mathbf{x}}'_n = \mathop{\arg\min}\limits_{\mathbf{x}'_n\in \mathbb{R}^M} \mathcal{L}\left(\mathbf{x}'_{n}, y'_{n}, \mathbf{x}_{n}, \lambda\right).\\
\end{aligned}
\end{equation}

There are many options for defining the distance metric $d\left(\mathbf{x}_{n}, \mathbf{x}'_{n}\right)$, such as Manhattan distance and Euclidean distance. Here, we present a simple measurement to estimate the distance between factual sample and counterfactual sample as follows
\begin{equation}
d\left(\mathbf{x}_{n}, \mathbf{x}'_{n}\right)=\sum_{m=1}^{M} \frac{\left|X_{nm}-X_{nm}'\right|}{\mathrm{MAD}_{m}},
\label{eq:objective}
\end{equation}
where $\mathrm{MAD}_{m}$ is the median absolute deviation of the $m$-th feature variable, i.e. $\mathrm{MAD}_{m}=\mathrm{median}(|X_{nm}-\mathrm{median}(X_{1:N,m}) |)$. It is robust to sample variance in the dataset. Also, it can refer to the population parameter that is estimated by the MAD calculated from a sample.

It is worth to mention that given a certain factual sample, our approach may generate more than one counterfactual samples according to Eq.(\ref{eq:objective}).
In our work, the only one with the minimal distance is selected as the counterfactual sample due to the requirement of ``minimum inversion'', resulting in that most of the new counterfactual samples concentrate near the decision boundary.

\subsection{Perturbation function}
Now it is straightforward to solve the objective in Eq.(\ref{eq:objective}) by an approximate optimizer to search in the feature space. However, this faces the awkward situation where computation cost is expensive, and would end up being intractable with the growth of the feature dimension.
To this end, we use a regular perturbation on each factual sample to generate counterfactual.
To simplify the objective, we first consider to generate counterfactual samples by perturbing only one feature.
We will give the complete algorithm about perturbing all the features in the later section.
Suppose we seek a perturbation on the $m$-th feature that can fool the classifier $f_{\widehat{\mathbf{w}}}$, i.e.,
\begin{equation}
f_{\widehat{\mathbf{w}}}\left(\mathbf{x}_{n}\right) \neq f_{\widehat{\mathbf{w}}}\left(\mathbf{x}'_{n}=\mathbf{x}_{n}+\Delta \mathbf{x}_n\right),
\label{eq:inversion}
\end{equation}
where $\Delta \mathbf{x}_n=[0,\ldots, \Delta X_{nm},\ldots, 0]$ ($\Delta X_{nm}\neq 0$) and $\Delta X_{nm}$ is the perturbation value of the $m$-th feature on the $n$-th sample.
That is to say, the classifier can correctly group $\mathbf{x}_{n}$ into the majority class $j$ (i.e., $\mathbf{x}_{n}\in \mathbf{X}_{j}$ and $f_{\widehat{\mathbf{w}}}\left(\mathbf{x}_{n}\right)=j$) but will treat $\mathbf{x}'_{n}$, which has a small change on $\mathbf{x}_n$,  as a sample of the minority class $i$ (i.e., $f_{\widehat{\mathbf{w}}}\left(\mathbf{x}'_{n}\right)=i$).

\begin{figure}[htbp]
\centerline{\includegraphics[width=0.95\columnwidth]{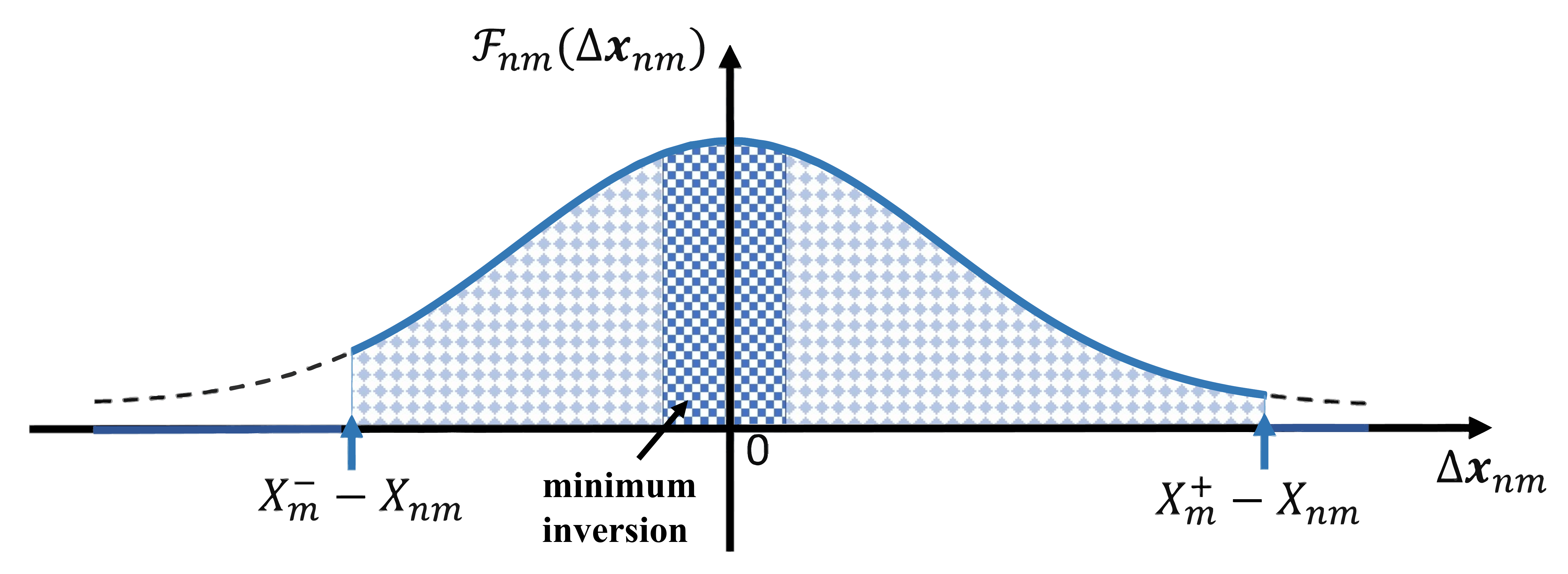}}
\caption{The truncated normal distribution of perturbation on factual samples.}
\label{fig:normaldistribution}
\end{figure}

To achieve such a requirement, we further introduce the truncated normal distribution $\mathcal{F}(\Delta\mathbf{x}_n)$, which is the probability distribution derived from that of a normally distributed random variable by bounding the generative counterfactual samples from both of below and above.
We develop a random perturbation procedure based on truncated normal distribution to perform an approximate learning for sampling the counterfactual samples.
More specifically, for any $m$-th feature of the factual sample $\mathbf{x}_n$, we estimate the distribution on the perturbation $\Delta\mathbf{x}_{nm}$ based on the following conditional probabilities shown as
\begin{equation}
\begin{split}
&\mathcal{F}_{nm}(\Delta \mathbf{x}_{nm} | X_{nm}, X^{-}_{m}, X^{+}_{m}, \sigma)\\
&= \left\{
    \begin{array}{l l}
        \frac{\frac{1}{\sigma}\psi\left(\frac{\Delta \mathbf{x}_{nm}}{\sigma}\right)}{\Phi\left(\frac{X^{+}_{m}-X_{nm}}{\sigma}\right)-\Phi\left(\frac{X^{-}_{m}-X_{nm}}{\sigma}\right)} & \text{\scriptsize{if $X^{-}_{m}\leq X_{nm}+\Delta\mathbf{x}_{nm}\leq X^{+}_{m}$}},\\
        0 & \text{otherwise},
    \end{array} \right.
\end{split}
\end{equation}
where $X^{+}_{m}$ and $X^{-}_{m}$ represent the maximum and minimum values of the $m$-th feature in the factual dataset $\mathbf{X}$, respectively, and $\sigma$ is the standard deviation of the $m$-th feature.
$\psi(z)=\frac{1}{\sqrt{2 \pi}} e^{-\frac{1}{2} z^{2}}$ is the probability density function of the standard normal distribution and $\Phi(z)$ is the cumulative distribution function as defined as follows
\begin{equation}
\begin{aligned}
\Phi(z)&=\frac{1}{2}\left(1+\operatorname{erf}\left(\frac{z}{\sqrt{2}}\right)\right)\\
&=\int_{-\infty}^{z} \frac{1}{\sqrt{2 \pi}} e^{-\frac{-t^{2}}{2}} d t,
\end{aligned}
\end{equation}
where $\operatorname{erf}\left(\cdot\right)$ is the Gaussian error function, which is a non-primary function.
In this way, a counterfactual sample of minority class can be generated by leveraging the rich information inherent in the majority class. Besides, since the probability mass concentration for $\mathcal{F}$ is greater near the point zero, as shown in Fig.~\ref{fig:normaldistribution}, there is a higher probability of generating the boundary sample close to $\mathbf{x}_n$, which satisfies the requirement of ``minimum inversion''. In addition, any $m$-th feature in the new sample will not exceed the original range of $X_{1:N,m}$.

Now we need to generate the concrete perturbation values $\Delta \mathbf{x}_{nm}$ that follow the distribution $\Delta \mathbf{x}_{nm}\sim\mathcal{F}_{nm}$ truncated to the range $[X^{-}_{m}-X_{nm}, X^{+}_{m}-X_{nm}]$.
One of the simplest way for implementing such sampling is the inverse transform method where the perturbation is defined as $\Delta \mathbf{x}_{nm}=\Phi^{-1}(\Phi(\alpha)+U\cdot(\Phi(\beta)-\Phi(\alpha)))\sigma+X_{nm}$ with  $\alpha=\frac{X^{+}_{m}-X_{nm}}{\sigma}$, $\beta=\frac{X^{-}_{m}-X_{nm}}{\sigma}$ and $U$ a uniform random number on $(0,1)$.
Although one of the simplest to perform an exact sampling, it is unfortunately that this method can either fail when sampling in the tail of the normal distribution, or be much too slow~\cite{Botev2017Simulation}.
Instead, in practice we utilize an alternative method of approximate sampling within a Gibbs sampling framework by introducing one latent variable~\cite{Damien2001Sampling}. It is more computationally efficient within a Gibbs sampling framework.

Finally, for any factual sample $\mathbf{x}_n$, the set of perturbations on the $m$-th feature can be constructed as follows
\begin{equation}
\begin{aligned}
\mathcal{C}_{nm}=&\{\Delta \mathbf{x}_n| \Delta\mathbf{x}_{nm}\sim\mathcal{F}_{nm}, \mathbf{x}'_{n}=\mathbf{x}_n+\Delta \mathbf{x}_n, \text{ and}\\
& d(\mathbf{x}_{n},\mathbf{x}'_{n})<\epsilon,\mathbf{x}_n\in \mathbf{X}_j, f_{\widehat{\mathbf{w}}}\left(\mathbf{x}_{n}\right)=j, f_{\widehat{\mathbf{w}}}\left(\mathbf{x}'_{n}\right)=i\},
\end{aligned}
\label{eq:perturbation}
\end{equation}
where $\epsilon$ is a constant that limits the perturbation to a certain magnitude.

\subsection{Algorithm and complexity analysis}
So far we have described to generate counterfactual samples by perturbing one feature.
It may potentially lead to loss of information. For example, only perturbing a certain feature may not find any counterfactual sample that satisfies the inversion requirement in Eq.(\ref{eq:inversion}), resulting in $\mathcal{C}_{nm}=\phi$.
As a remedy of this issue, we extend the model by perturbing any combination of features as follows
\begin{equation}
\begin{aligned}
\mathcal{C}_{n}&=\{\Delta \mathbf{x}_n| \text{for any $m$}: \Delta\mathbf{x}_{nm}\sim\mathcal{F}_{nm} \text{ or } \Delta\mathbf{x}_{nm}=0,\\
&  \text{and } d(\mathbf{x}_{n},\mathbf{x}'_{n})<\epsilon,\mathbf{x}_n\in \mathbf{X}_j, f_{\widehat{\mathbf{w}}}\left(\mathbf{x}_{n}\right)=j, f_{\widehat{\mathbf{w}}}\left(\mathbf{x}'_{n}\right)=i\}.
\end{aligned}
\label{eq:extended}
\end{equation}

As there are a number of $M$ features, it will totally produce $M!$ combinations (Eq.(\ref{eq:extended})), which is unacceptable in practice.
To reduce the complexity, we first estimate the feature importance, and then perturb the ranked features in order.
Here we adopt the Spearman correlation algorithm~\cite{Sedgwick2014Spearman} to implement the feature ranking.
Without loss of generality, we assume that the features are ranked in terms of their importance as $<F_1, \ldots, F_M>$.
Our perturbation will be conducted in ${M}$ rounds. For each $m$ where $m=1\ldots M$, $F_m$ is perturbed ${T}$ times.
In detail, in the first round the set of perturbations on the most important feature $F_1$ is constructed according to Eq.(\ref{eq:perturbation}), denoted by $\mathcal{C}_{n,1}$.
Generally, in the $m$-th round, the set of perturbations on the first $m$ most important features $F_1,\ldots, F_m$ can be constructed as follows
\begin{equation}
\begin{aligned}
\mathcal{C}_{n, 1:m}&=\{\Delta \mathbf{x}_n| \text{for $k=[1:m]$}: \Delta\mathbf{x}_{nk}\sim\mathcal{F}_{nk}, \text{and }\\
&d(\mathbf{x}_{n},\mathbf{x}'_{n})<\epsilon,\mathbf{x}_n\in \mathbf{X}_j, f_{\widehat{\mathbf{w}}}\left(\mathbf{x}_{n}\right)=j, f_{\widehat{\mathbf{w}}}\left(\mathbf{x}'_{n}\right)=i\}.
\end{aligned}
\label{eq:optimizedperburtation}
\end{equation}
After $M$ rounds, the set of perturbations can be constructed by combining all the sets generated in each round as follows
\begin{equation}
\mathcal{C}_{n}=\mathcal{C}_{n,1}\cup\ldots\cup \mathcal{C}_{n,1:m} \cup\ldots \cup \mathcal{C}_{n, 1:M}.
\label{eq:finalperburtation}
\end{equation}
In this way, the number of combination is enormously reduced to $M$, which may generate counterfactual samples more efficiently.
At the end we can find an optimal counterfactual sample $\widehat{\mathbf{x}}'_n=\mathbf{x}_n+\Delta\mathbf{x}_n$ (Eq.(\ref{eq:objective})) with the minimal distance $d(\mathbf{x}_n,\widehat{\mathbf{x}}'_n)$ such that $\Delta \mathbf{x}_n\in\mathcal{C}_{n}$.
The complete procedure of our counterfactual-based oversampling algorithm is given in Algorithm~\ref{algo-gp}. Theoretically, for any pair of imbalanced classes, its time complexity is $\mathcal{O}(NM^2T)$.
\begin{algorithm*}
  \caption{The Counterfactual-Based Oversampling}\label{algo-gp}
  \small
  \begin{algorithmic}[1]
    \Procedure{Generating\_Counterfactual\_Samples}{$\mathbf{X},i,j$} \Comment{For any pair of imbalanced classes $i,j$ where $N_i<N_j$}
     \State Train a classifier $f_{\widehat{\mathbf{w}}}$ over the dataset $\mathbf{X}$;
     \State Rank the feature importance $\mathcal{I}_F=<F_1, \ldots, F_M>$;
     \For{each sample $\mathbf{x}_n\in\mathbf{X}_j$ }
         \For{$m=1$ to $M$} \Comment{Perturb features in order according to $\mathcal{I}_F$}
            \For{$t=1$ to $T$} \Comment{Perturb $T$ times for each sample}
              \State $\Delta \mathbf{x}_n = \{\Delta\mathbf{x}_{nk}\sim\mathcal{F}_{nk}, k=[1:m]\}$; \Comment{Perturb the first $m$ features by sampling over $\mathcal{F}_{nk}$}
              \State $\mathbf{x}'_n=\mathbf{x}_n+\Delta\mathbf{x}_n$;
              \If{$d(\mathbf{x}_{n},\mathbf{x}'_{n})<\epsilon$ and $f_{\widehat{\mathbf{w}}}\left(\mathbf{x}_{n}\right)=j$ and $f_{\widehat{\mathbf{w}}}\left(\mathbf{x}'_{n}\right)=i$} \Comment{Check the inversion of $\mathbf{x}_n$ with $f_{\widehat{\mathbf{w}}}$ (Eq.(\ref{eq:optimizedperburtation}))}
                \State $\mathcal{C}_{n,1:m}\leftarrow\mathcal{C}_{n,1:m}\cup \{\Delta\mathbf{x}_n\}$;
              \EndIf
            \EndFor
         \EndFor
      \State $\mathcal{C}_{n}=\mathcal{C}_{n,1}\cup\ldots\cup \mathcal{C}_{n,1:m} \cup\ldots \cup \mathcal{C}_{n, 1:M}$; \Comment{Combine the perturbation sets (Eq.(\ref{eq:finalperburtation})) }
      \State $\widehat{\mathbf{x}}'_n = {\arg\min}_{\Delta\mathbf{x}_n\in\mathcal{C}_{n}}d\left(\mathbf{x}_{n}, \mathbf{x}'_{n}\right)$; \Comment{Choose the counterfactual sample with minimal inversion}
      \State $\mathbf{X}_i\leftarrow \mathbf{X}_i\cup\{\widehat{\mathbf{x}}'_{n}\}$;
     \EndFor
     \State \textbf{return} $\mathbf{X}_i$;
    \EndProcedure
\end{algorithmic}
\end{algorithm*}

\section{Empirical evaluations}

\subsection{Datasets}
We have considered five datasets, including one synthetic dataset and four real-world datasets in our experiments.\\
\textbf{Synthetic dataset}. Our synthetic dataset is a binary class dataset with a total number of $1,000$ samples, including only $83$ minority samples. These samples contain only two features, whose values are randomly sampled normal distributions. In addition, we intendedly placed a small number of noise points, which are minority samples close to and even surrounded by the majority samples.

The four real-world datasets include three publicly-available benchmark datasets and one in-house dataset which we collected from a local hospital.
These datasets involve both binary and multiple classes, and contain various data types such as tabular data and image data.\\
\textbf{Wisconsin breast cancer (WBC) dataset}~\cite{UCI2007}. This publicly-available dataset mainly shows the relationship between nine features of cancer cell influence factors and two cancer risk classes (benign or malignant). It contains $699$ samples with an imbalance ratio $1.9:1$.\\
\textbf{Fashion-MNIST (FMN) dataset}~\cite{articlemnist}. This dataset is an image dataset containing $10$ classes. We chose two of them and constructed an imbalance dataset of $3,190$ samples by manually selection. The imbalance ratio is $14.95:1$.\\
\textbf{Page blocks classification (PBC) dataset}~\cite{UCI2007}. This dataset is a multi-class dataset with five classes of the page layouts of documents. It consists of $5,474$ samples with $10$ features and imbalance ratios are $175.464:11.750:4.107:3.143:1$.\\
\textbf{In-house lung cancer (ILC) dataset}. This dataset is collected from a hospital through questionnaires and expert diagnosis. There are $120$ features that are mainly cataloged into seven aspects: basic information of patients, dietary habits, living environment, routines and habits, psychology and emotions, past and family history of the disease, and female physiology and fertility. The $12,330$ samples in this dataset have the imbalance ratio $4.366:1$. 

These four datasets contain unique challenges:
The WBC dataset is comprised of a handful of samples with scarce features; the ILC dataset involves a large number of features;
the FMN dataset contain image data in highly imbalanced classes; and
the PBC dataset has multiple classes with extreme imbalance ratios among them.

\subsection{Experiment setup}
We conducted the following two experiments. Firstly, we started by evaluating the generation of boundary samples and comparing our approach with two existing oversampling approaches, SSO~\cite{Rong2014Stochastic} and ADASYN~\cite{He2008ADASYN} (\emph{Experiment I} in Section~\ref{sec:exp1}).
Secondly,we compared the classification performance of our approach with three commonly used resampling methods, SOMO~\cite{douzas2017self}, k-SMOTE~\cite{Georgios2018Improving} and CUSBoost~\cite{RayhanCUSBoost} (\emph{Experiment II} in Section~\ref{sec:exp2}).
The first two are oversampling methods, while the last one is undersampling.
Our evaluation metrics are \emph{F-measure} and \emph{G-Mean}.
In detail, \emph{F-measure} is the harmonic mean of the precision and recall, measuring the test accuracy.
\emph{G-Mean} is the geometric mean, also called Fowlkes-Mallows index, which is an effective indicator to detect classification performance when the training data is imbalanced distributed. Formally, \emph{G-Mean} is defined as follows:
\begin{equation}
\text{\emph{G-Mean}}=\sqrt{TPR \times TNR},
\end{equation}
where $TPR$ and $TNR$ denote the true positive rate and the true negative rate, respectively.
We utilized six common classifiers in our experiments, namely, SVM, KNN, NN, GBDT, RF, and LR. Since some of those classifiers (i.e. RF, NN) are probabilistic models with uncertain results, the evaluation result is the average value of $100$ runs.


\subsection{Experiment I: Evaluation of boundary sample generation on synthetic dataset}
\label{sec:exp1}
\begin{figure*}[htbp]
\centering
\subfigure[Original dataset]{
\includegraphics[width=0.35\textwidth]{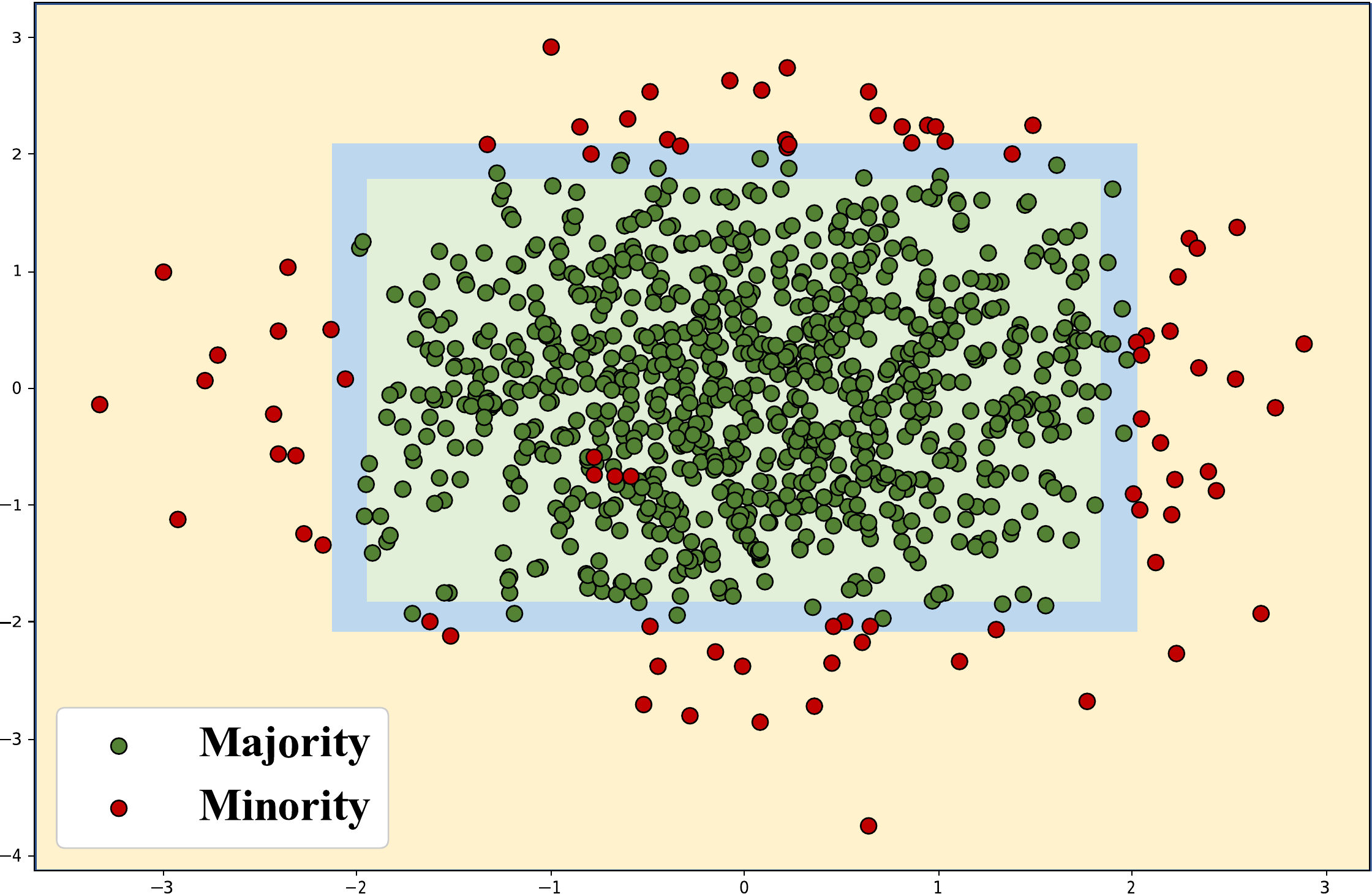}
}
\subfigure[SSO]{
\includegraphics[width=0.35\textwidth]{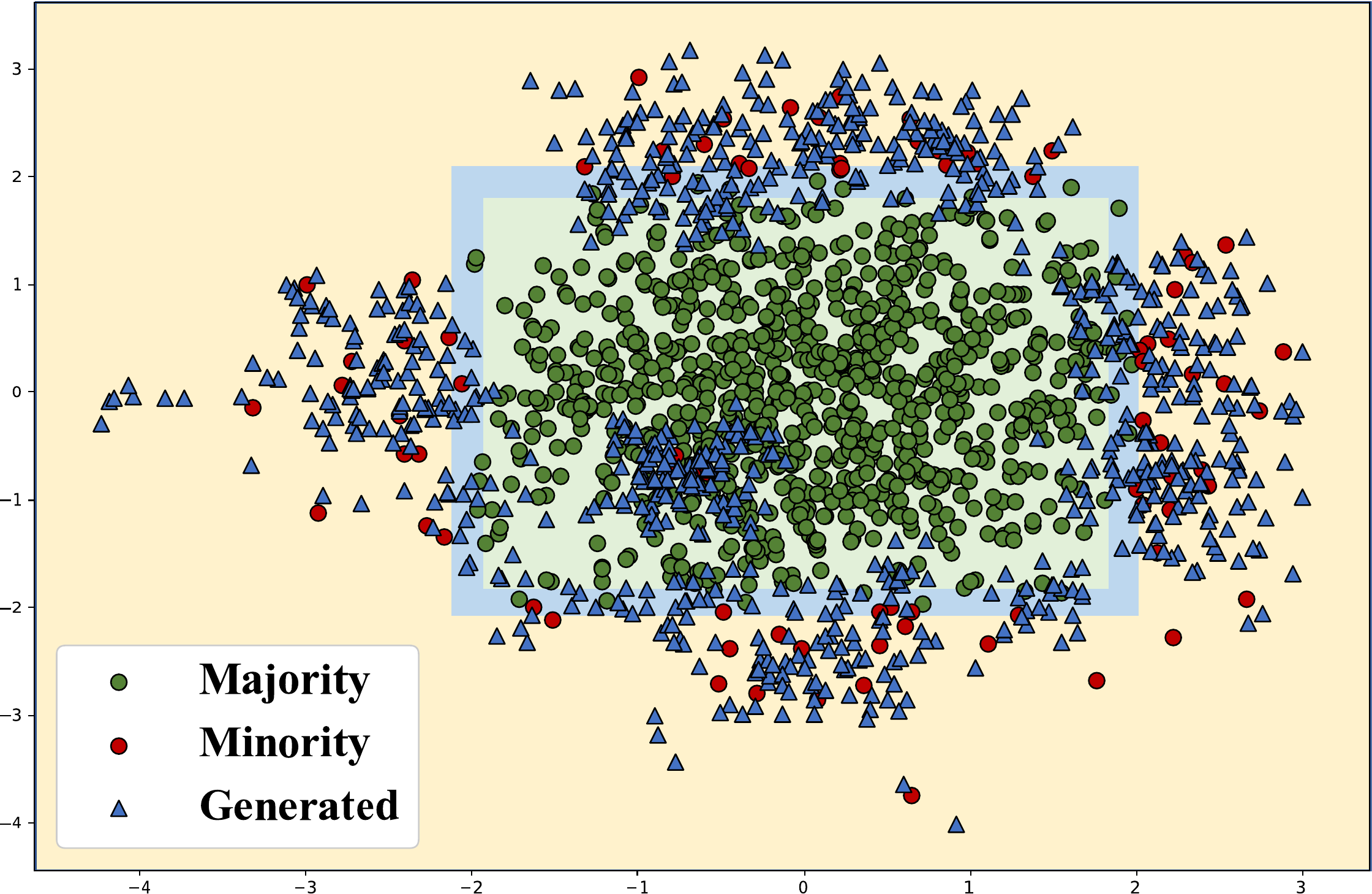}
}
\subfigure[ADASYN]{
\includegraphics[width=0.35\textwidth]{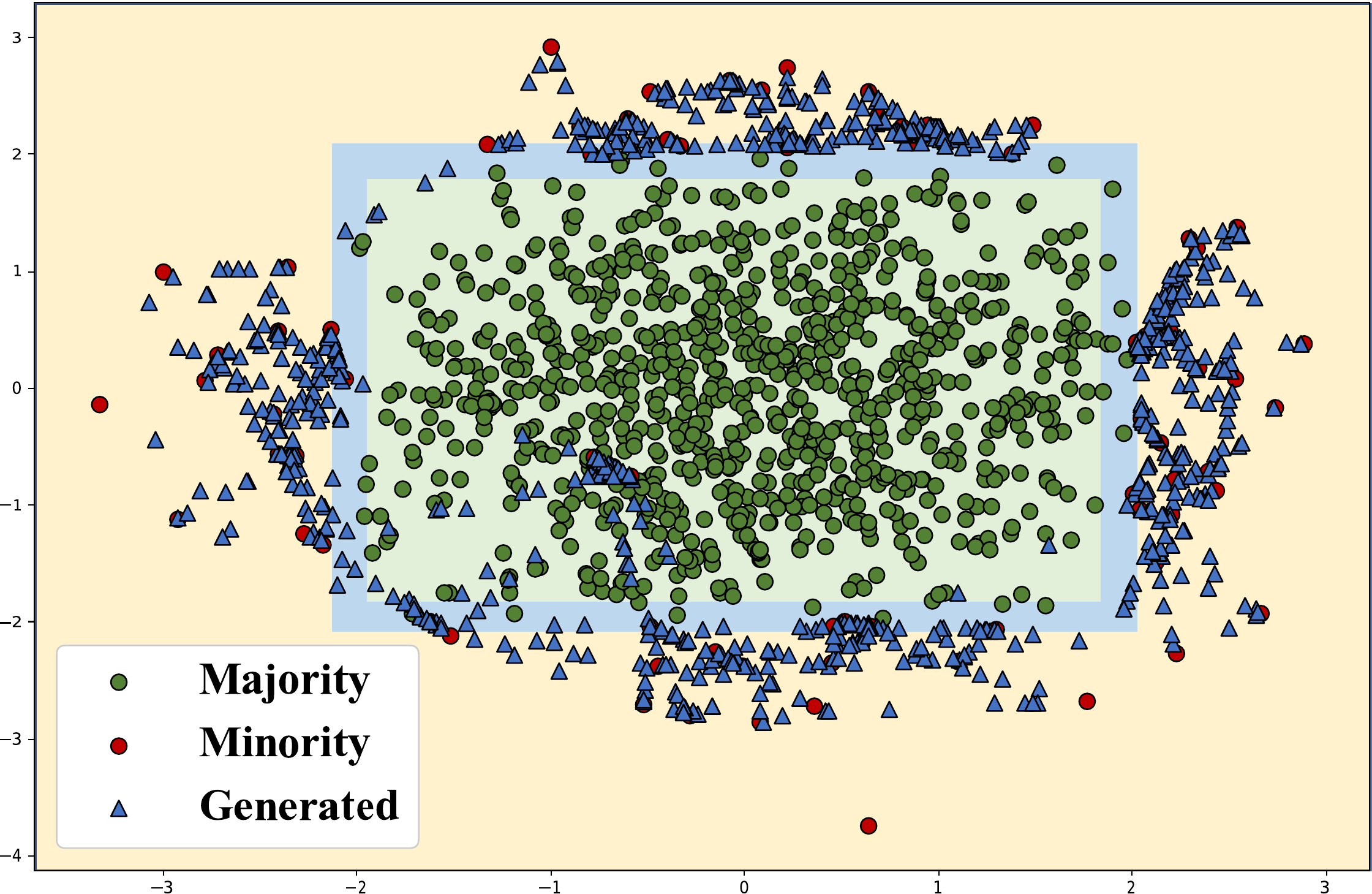}
}
\subfigure[Our approach]{
\includegraphics[width=0.35\textwidth]{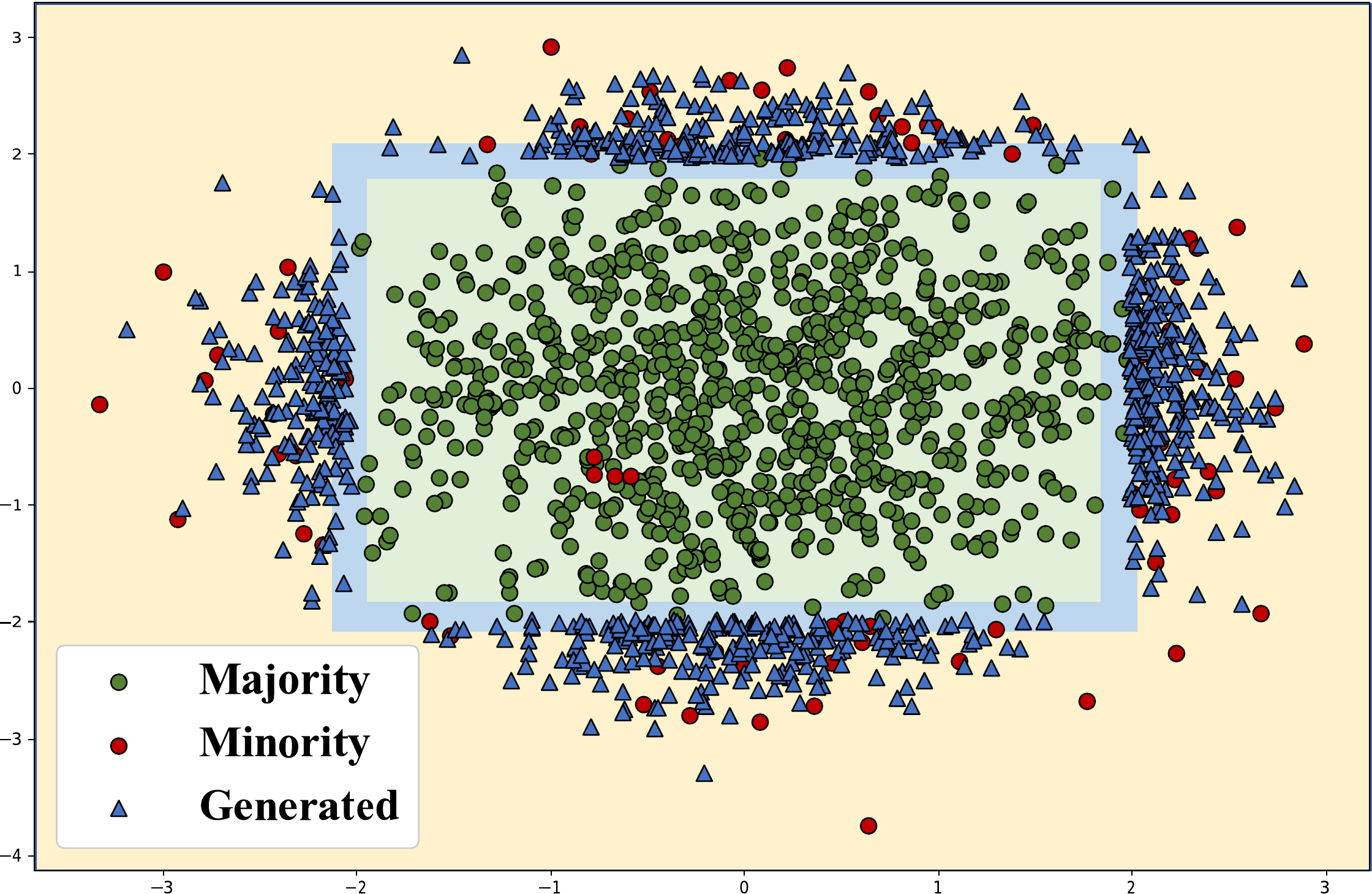}
}
\caption{Plot of features from the majority samples (green dots) and minority samples (red dots) over the synthetic dataset. (a) shows the distribution of the original samples, while (b)(c)(d) shows the samples after oversampling by the respective methods, where blue triangles represent the new samples. The feature space is divide into three parts:
\emph{boundary minority} space that is near the boundary (blue area), \emph{majority} space (green area) and \emph{interior minority} space that is far from the boundary (yellow area).}
\label{fig:generate}
\end{figure*}
Fig.~\ref{fig:generate} visually illustrates the generation of minority samples by using the three comparative oversampling methods over the synthetic dataset. It can be seen that these methods generate approximately the same number of new minority samples.
It is obvious that SSO generates a large amount of minority samples distributed randomly in the feature space, but most of these new samples are meaningless to enhance the decision boundary. Even worse, some of them are noise points locating in the majority area.
ADASYN can generate more samples near the decision boundary than SSO, but it still creates a number of noise points.
It is clear that our approach outperforms the other two oversampling methods, as most of the new samples are close to the decision boundary but do not constitute noise points. This is because our counterfactual approach inverts majority samples to minority classes in a minimal distance.

Table~\ref{tab-generation} reports the statistical results of the counterfactual sample distributions.
For simplicity, these counterfactual samples are categorized into three groups: those in the minority area and near the boundary (\emph{boundary minority}), those in the minority area but far from the boundary (\emph{interior minority}), and those in the majority area (\emph{majority}).
Our approach significantly outperforms SSO and ADASYN in the boundary sample generation with a large margin. Unlike the other two methods, our approach generates counterfactual samples by perturbations only and thus guarantees the locations of those samples in the minority area.
Notably, our approach satisfies the ``minimum inversion'' in the counterfactual theory. This explains the reason that our approach can generate a large proportion of boundary samples and therefore enhance the decision boundary of a classifier.

\begin{table}[htbp]
\centering
\caption{Comparison on the proportions of the generated sample distributions.}\label{tab-generation}
\scalebox{0.76}[0.76]{
\begin{tabular}{@{}ccccccccccc@{}}
\toprule
\multirow{1}{*}              & \multirow{2}{*}{\textbf{Method}} & \multicolumn{2}{c}{\textbf{Majority}} & \multicolumn{2}{c}{\textbf{Boundary minority} } & \multicolumn{2}{c}{\textbf{Interior minority}} & \\ \cmidrule(l){3-4}\cmidrule(l){5-6}\cmidrule(l){7-8}
                                                       &                         & original & generated & original & generated & original & generated      & \\ \midrule
\multirow{3}{*}                                         & \textbf{SSO}           & 4 & 265(31.9\%) & 19 & 108(13.0\%) & 60 & \textbf{457(55.1\%)}  & \\
                                                       & \textbf{ADASYN}       & 4 & 90(10.6\%) & 19 & 182(21.4\%) & 60 & 580(68.1\%) & \\
                                                       & \textbf{Ours}           & 4 & \textbf{0(0\%)} & 19 & \textbf{314(37.5\%)} & 60 & 523(62.5\%) & \\ \bottomrule
\end{tabular}
}
\end{table}

\subsection{Experiment II: Evaluation on real-world datasets}
\label{sec:exp2}
\begin{figure}[htbp]
\centering
\subfigure[WBC]{
\includegraphics[width=0.45\columnwidth]{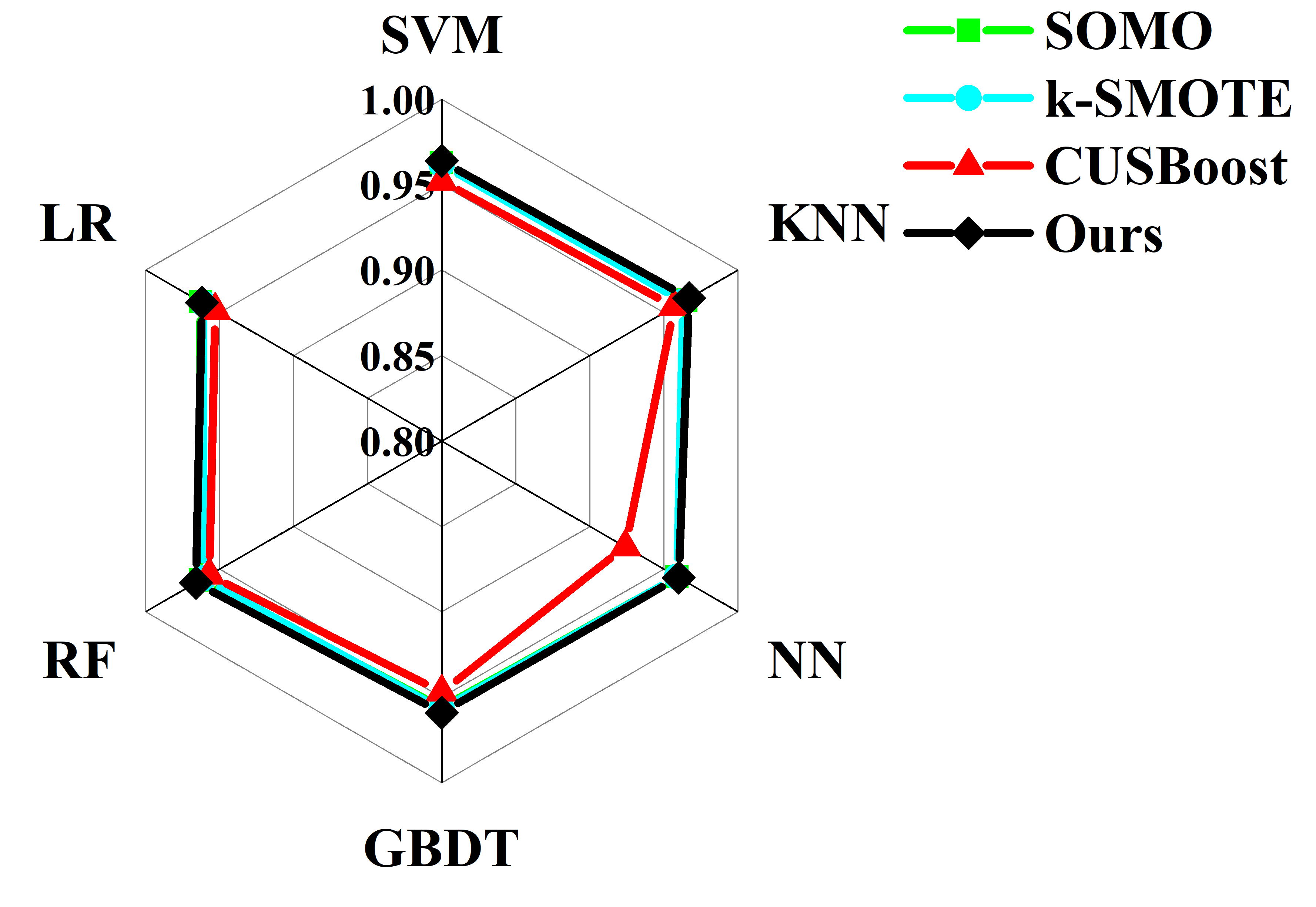}}
\subfigure[ILC]{
\includegraphics[width=0.45\columnwidth]{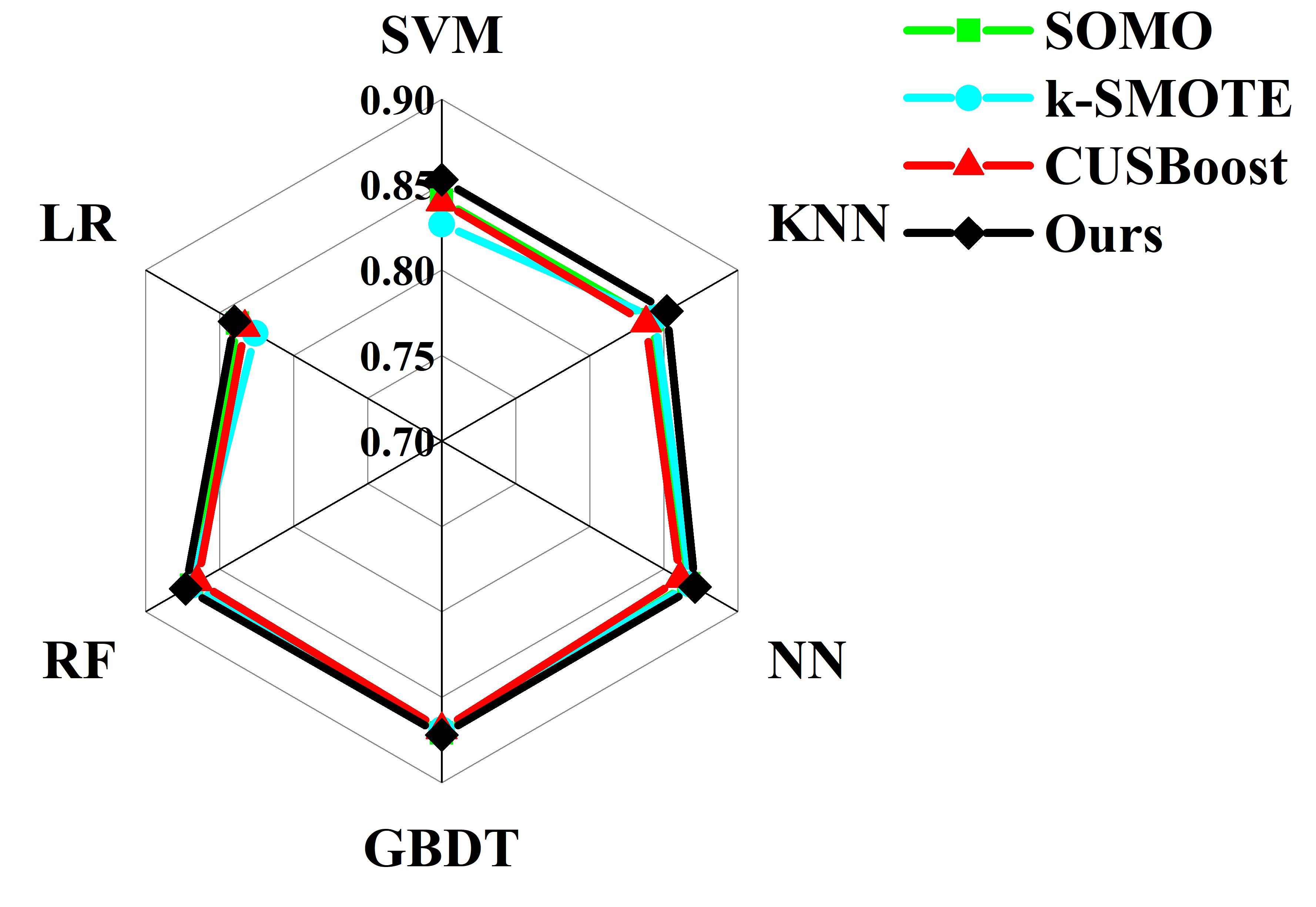}}
\subfigure[FMN]{
\includegraphics[width=0.45\columnwidth]{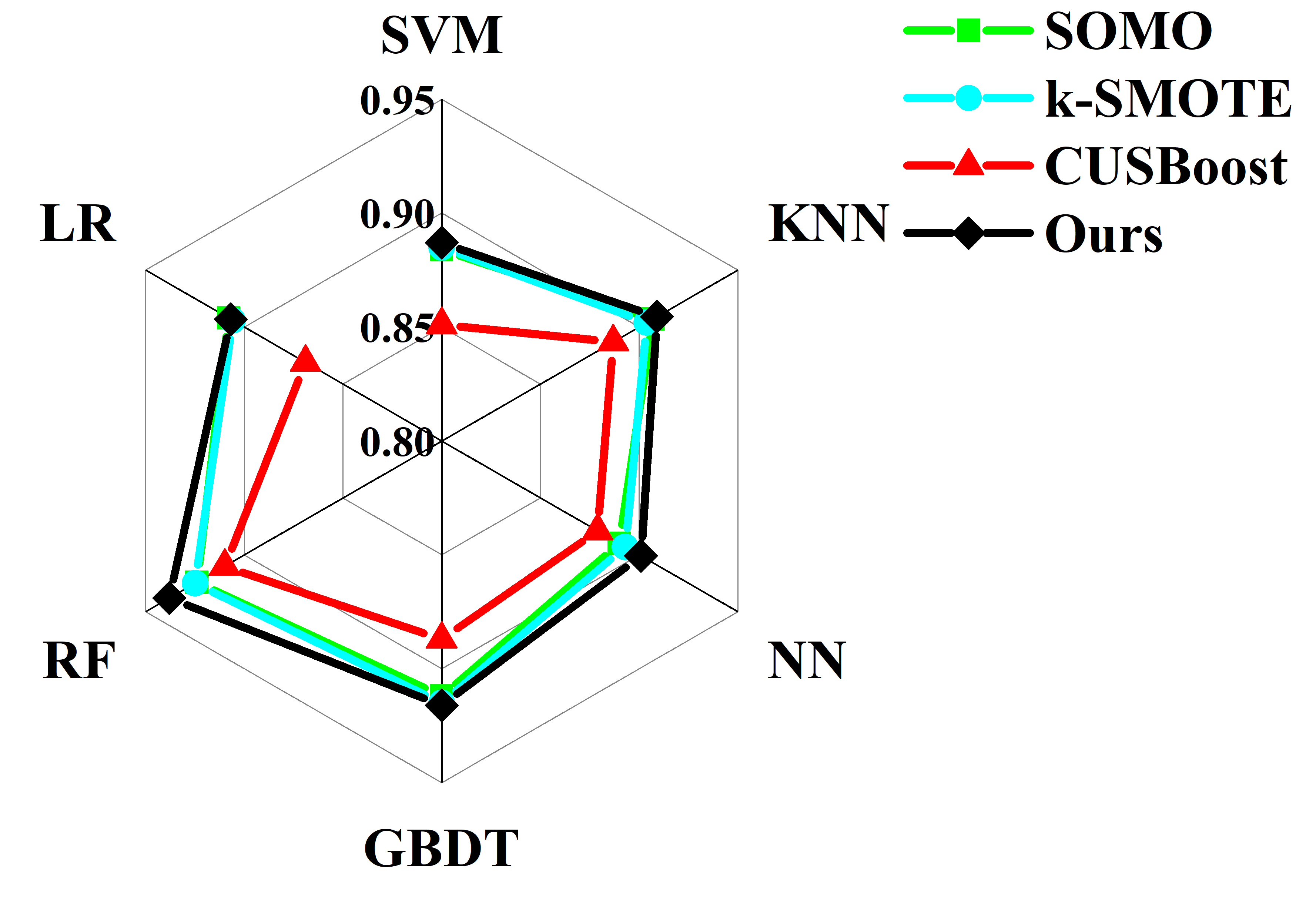}}
\subfigure[PBC]{
\label{fig:experiment1:d}
\includegraphics[width=0.45\columnwidth]{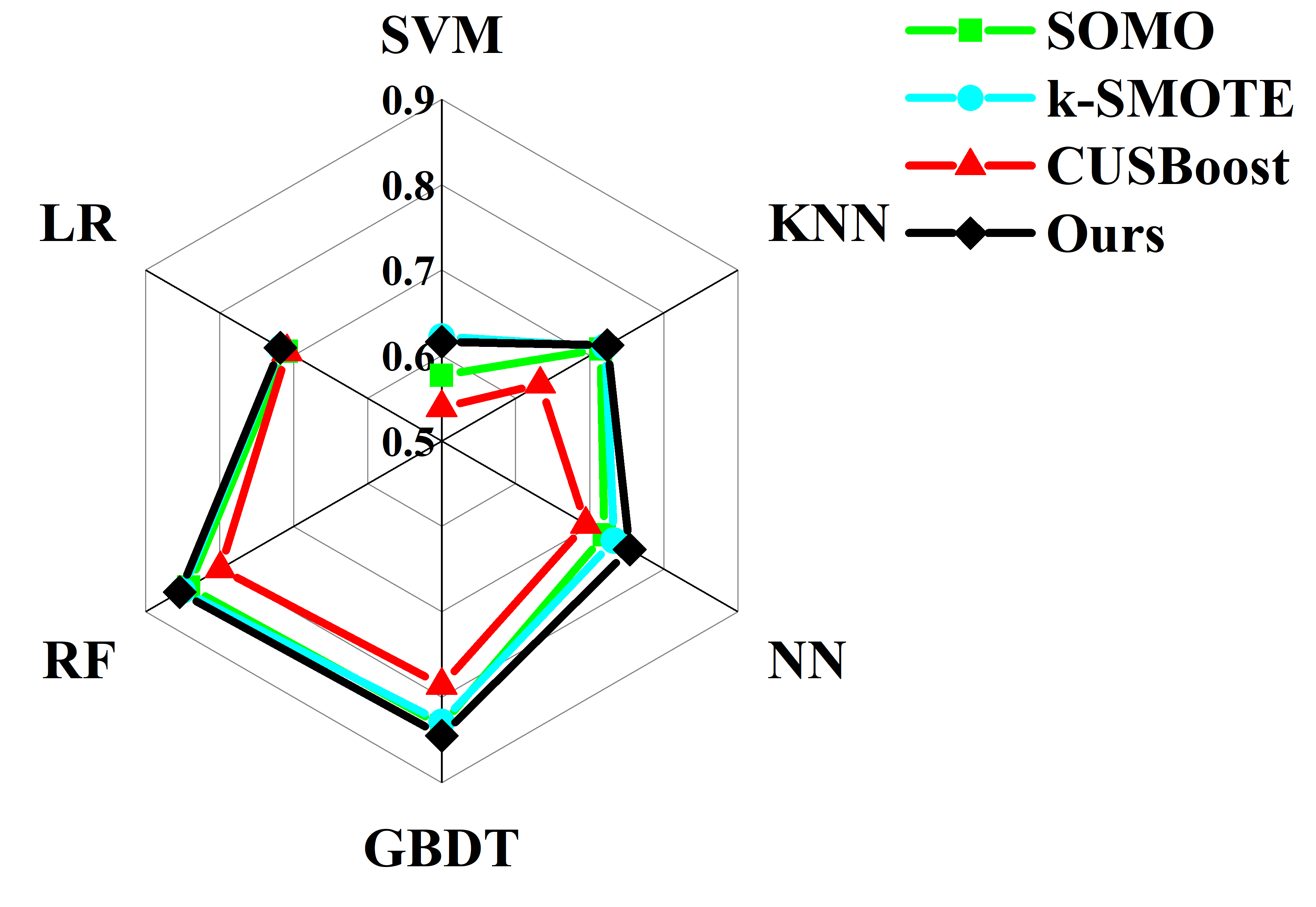}}
\caption{\emph{F-measure} comparison against other resampling methods.}
\label{fig:experiment1}
\end{figure}

\textbf{Comparison of classification performance against other methods.}
Figure~\ref{fig:experiment1} shows the averaged \emph{F-measure} results over 10-fold cross-validations. Our approach outperforms the other methods substantially on all four datasets.
This mainly due to its abilities to take advantage of the majority samples with rich information inherent in them.
Apparently, our approach performs stably for all six classifiers. It indicates that the samples generated by our approach are more classifier-agnostic than those of other methods and can be adopted by different classifiers.

\textbf{\emph{G-Mean} comparison.} \emph{G-Mean} is commonly used for assessing the performance under imbalanced domains. It computes the geometric mean of accuracies of the two classes.
Table~\ref{tab-dataset2} depicts the comparison results among resampling methods.
The oversampling methods (SOMO, k-SMOTE and ours) are more effective to imbalanced classification than the undersampling method (CUSBoost) as being capable to generate more minority samples.
Moreover, it is clear that our method is significantly more accurate (about $5\%-15\%$) than other resampling methods.
This is mainly because the other resampling methods are greatly affected by noise points and interior minority points, which often appear in imbalanced datasets.
Those methods only take minority samples or neighbouring samples into consideration but ignore the rich information in the majority classes, leading to be overfitted to the minority classes.
This explains why our approach achieves the best performance (with nearly $15\%$ boost at most compared with CUSBoost) among all compared methods for the PBC dataset, which includes extremely imbalanced samples in multiple classes.

\begin{table}[]
\centering
\caption{\small{\emph{G-Mean} comparison against other resampling methods. The value in brackets is the standard deviation.}}\label{tab-dataset2}
\scalebox{0.55}[0.55]{
\begin{tabular}{@{}lllllllll@{}}
\toprule
\multirow{2}{*}{\textbf{\quad Dataset}}               & \multirow{2}{*}{\textbf{Method}} & \multicolumn{6}{c}{\textbf{Classifiers} }                 & \\ \cmidrule(l){3-9}
                                                       &                         & SVM & KNN & NN & GBDT & RF & LR          & \\ \midrule
\multirow{4}{*}{\quad \textbf{WBC}}                    & \textbf{SOMO}              & 0.964($\pm$0.012) & 0.968($\pm$0.012) & 0.959($\pm$0.014) & 0.957($\pm$0.016) & 0.966($\pm$0.013) & 0.964($\pm$0.013)    & \\
                                                       & \textbf{k-SMOTE}                      & 0.963($\pm$0.012) & 0.969($\pm$0.013) & 0.959($\pm$0.014) & 0.959($\pm$0.015) & 0.965($\pm$0.013) & 0.962($\pm$0.014) & \\
                                                       & \textbf{CUSBoost}                    & 0.951($\pm$0.018) & 0.955($\pm$0.016) & 0.918($\pm$0.024) & 0.945($\pm$0.018) & 0.958($\pm$0.017) & 0.95($\pm$0.02)
                                                       \\
                                                       & \textbf{Ours}                   & \textbf{0.968($\pm$0.011)} & \textbf{0.97($\pm$0.01)} & \textbf{0.963($\pm$0.012)} & \textbf{0.964($\pm$0.015)}& \textbf{0.967($\pm$0.011)} & \textbf{0.965($\pm$0.01)}  & \\ \midrule
\multirow{4}{*}{\quad \textbf{ILC}}                     & \textbf{SOMO}              & 0.839($\pm$0.008) & 0.846($\pm$0.045) & 0.889($\pm$0.017) & 0.894($\pm$0.008) & 0.892($\pm$0.008) & 0.836($\pm$0.008)   & \\
                                                       & \textbf{k-SMOTE}                      & 0.826($\pm$0.034) & 0.854($\pm$0.041) & 0.892($\pm$0.021) & 0.891($\pm$0.01) & 0.893($\pm$0.009) & 0.822($\pm$0.035) & \\
                                                       & \textbf{CUSBoost}                   & 0.839($\pm$0.013) & 0.842($\pm$0.041) & 0.875($\pm$0.026) & 0.889($\pm$0.008) & 0.885($\pm$0.012) & 0.824($\pm$0.031) & \\
                                                       & \textbf{Ours}                    &
                                                       \textbf{0.845($\pm$0.007)} &
                                                       \textbf{0.861($\pm$0.043)} & \textbf{0.898($\pm$0.018)} & \textbf{0.9($\pm$0.006)} & \textbf{0.895($\pm$0.009)} & \textbf{0.844($\pm$0.008)} & \\ \midrule
\multirow{4}{*}{\quad \textbf{FMN}}                    & \textbf{SOMO}              & 0.869($\pm$0.064) & 0.857($\pm$0.031) & 0.893($\pm$0.042) & 0.881($\pm$0.034) & 0.893($\pm$0.026) & 0.89($\pm$0.026)    & \\
                                                       & \textbf{k-SMOTE}                      & 0.89($\pm$0.058)  & 0.872($\pm$0.028) & 0.892($\pm$0.037) & \textbf{0.895($\pm$0.025)} & 0.897($\pm$0.024) & 0.902($\pm$0.028) & \\
                                                       & \textbf{CUSBoost}                     & 0.864($\pm$0.049) & 0.825($\pm$0.036) & 0.877($\pm$0.048) & 0.848($\pm$0.036)& 0.869($\pm$0.033) & 0.874($\pm$0.031) & \\
                                                       & \textbf{Ours}                    & \textbf{0.902($\pm$0.037)} & \textbf{0.874($\pm$0.031)} & \textbf{0.904($\pm$0.036)} & 0.886($\pm$0.024) & \textbf{0.911($\pm$0.025)} & \textbf{0.913($\pm$0.03)} & \\ \midrule
\multirow{4}{*}{\quad \textbf{PBC}}                    & \textbf{SOMO}              & 0.616($\pm$0.074) & 0.669($\pm$0.042) &  0.714($\pm$0.083) & 0.823($\pm$0.04) & 0.83($\pm$0.04) & 0.662($\pm$0.056)   & \\
                                                       & \textbf{k-SMOTE}                      & \textbf{0.653($\pm$0.075)} & 0.671($\pm$0.043) & 0.712($\pm$0.057) & 0.819($\pm$0.04) & 0.838($\pm$0.041) & 0.669($\pm$0.052) & \\
                                                       & \textbf{CUSBoost}                   & 0.598($\pm$0.072) & 0.583($\pm$0.059) & 0.691($\pm$0.075) & 0.769($\pm$0.055)& 0.772($\pm$0.055) & 0.662($\pm$0.063) & \\
                                                       & \textbf{Ours}                    & 0.642($\pm$0.063) & \textbf{0.691($\pm$0.036)} & \textbf{0.745($\pm$0.057)}
                                                       & \textbf{0.843($\pm$0.031)} & \textbf{0.842($\pm$0.037)} & \textbf{0.699($\pm$0.035)} & \\ \bottomrule
\end{tabular}
}
\end{table}

\textbf{Performance evaluation on multi-class imbalanced dataset.}
We evaluate the multiclassification performance of all compared methods on the PBC dataset with multiple classes of imbalanced samples.
Fig.~\ref{fig:experiment1:d} and Table~\ref{tab-dataset2} have already shown that our method outperforms other methods on the PBC dataset.
Under exactly the same settings of a certain classifier, the number of correctly classified samples for each class is further evaluated after resampling by different methods.
Table~\ref{tab-Correct-prediction} depicts the averaged results by using the GDBT classifier over $10$ runs.
Overall, our counterfactual-based method relatively outperforms other resampling methods as being capable to enhance the minority boundaries. More specifically, it is clear that our approach significantly improves the classification of the minorities.
It is interesting to observe that all resampling methods
expect CUSBoost perform slightly worse on the majority classification than that without resampling (Original). This may be due to the fact that oversampling increases the minority area while decreases the majority space. Subsequently, the generation of a large number of minority samples impairs the classification performance on majority class to a certain extent.
In this dataset the Minority\_3 class is critical for the entire classification performance as it only contains a small number of samples. Other oversampling methods such as SOMO and k-SMOTE, which primarily depend on the minority, cannot obtain enough information from the minority class to generate new samples.
\begin{table}[htbp]
\centering
\caption{Comparison on the number of correctly classified samples in each class (using GDBT classifier). The value in the bracket shows the difference taking the ground truth as a baseline.}\label{tab-Correct-prediction}
\scalebox{0.8}{
\begin{tabular}{@{}ccccccccccc@{}}
\toprule
\multirow{1}{*}              & \multirow{2}{*}{\textbf{Method}}  & \multirow{2}{*}{Majority}   & \multirow{2}{*}{Minority\_1}   & \multirow{2}{*}{Minority\_2}& \multirow{2}{*}{Minority\_3} & \multirow{2}{*}{Minority\_4} & \\
                                                       &        & \\
\midrule
\multirow{3}{*}                             & \textbf{Groud truth}   & 1792 & 43 & 121 & 10  & 32 & \\
\midrule
                                            & \textbf{Original}    & 1773(-19) & \textbf{27(-16)} & 107(-14) & 7(-3)  & \textbf{27(-5)} & \\
                                            & \textbf{SOMO}        & 1772(-20) & 26(-17) & 108(-13) & 7(-3)  & 26(-6) & \\
                                            & \textbf{k-SMOTE}    & 1771(-21) & \textbf{27(-16)} & 106(-15) & 7(-3)  & \textbf{27(-5)} & \\
                                            & \textbf{CUSBoost}   & \textbf{1774(-18)} & 23(-20) & 102(-19) & 6(-4)  & \textbf{27(-5)} & \\
                                            & \textbf{Ours}       & 1771(-21) & \textbf{27(-16)} & \textbf{111(-10)} & \textbf{8(-2)}  & \textbf{27(-5)} & \\
                                                        \bottomrule
\end{tabular}
}
\end{table}

\section{Conclusion and Future Work}
In this paper, we present a new counterfactual-based oversampling method where perturbations are incorporated into the majority samples in order to generate the minority samples. It can capture the rich inherent information of the majority class by minimum inversions on sample features. It is more reliable and flexible than existing methods for imbalanced classification.
As for future work, we plan to further relax the assumption that a classifier and its decision boundary are fixed and embed the learning of the classifier in the process of oversampling.


%
%
%



\end{document}